\documentclass{ieeeaccess}
\usepackage{cite}
\usepackage{amsmath,amssymb,amsfonts}
\usepackage{algorithmic}
\usepackage{graphicx}
\usepackage{textcomp}
\usepackage{xcolor}
\usepackage{upgreek}
\usepackage{soul}

\usepackage{hyperref}
\hypersetup{
    colorlinks=true,
    allcolors=blue
    }
\renewcommand{\figureautorefname}{Fig.}

\usepackage{multirow}
\usepackage{array}

\usepackage{bm}
\usepackage{makecell}

\usepackage{floatrow}
\usepackage{booktabs}
\usepackage[caption=false, subrefformat=parens, labelformat=parens]{subfig}
\usepackage{multirow}
\usepackage{array}
\newcommand{\PreserveBackslash}[1]{\let\temp=\\#1\let\\=\temp}
\newcolumntype{C}[1]{>{\PreserveBackslash\centering}p{#1}}
\newcolumntype{R}[1]{>{\PreserveBackslash\raggedleft}p{#1}}
\newcolumntype{L}[1]{>{\PreserveBackslash\raggedright}p{#1}}

\newcommand{\xmark}{$\times$}
\newcommand{\modelName}{FUSED-Net}

\makeatletter
\AtBeginDocument{\DeclareMathVersion{bold}
\SetSymbolFont{operators}{bold}{T1}{times}{b}{n}
\SetSymbolFont{NewLetters}{bold}{T1}{times}{b}{it}
\SetMathAlphabet{\mathrm}{bold}{T1}{times}{b}{n}
\SetMathAlphabet{\mathit}{bold}{T1}{times}{b}{it}
\SetMathAlphabet{\mathbf}{bold}{T1}{times}{b}{n}
\SetMathAlphabet{\mathtt}{bold}{OT1}{pcr}{b}{n}
\SetSymbolFont{symbols}{bold}{OMS}{cmsy}{b}{n}
\renewcommand\boldmath{\@nomath\boldmath\mathversion{bold}}}
\makeatother

\def\BibTeX{{\rm B\kern-.05em{\sc i\kern-.025em b}\kern-.08em
    T\kern-.1667em\lower.7ex\hbox{E}\kern-.125emX}}

\begin{document}
\history{Date of publication xxxx 00, 0000, date of current version xxxx 00, 0000.}
\doi{10.1109/ACCESS.2026.0429000}

\title{\modelName: Detecting Traffic Signs with Limited Data}
\author{
    \uppercase{Md. Atiqur Rahman}\authorrefmark{1}\authorrefmark{*},
    \uppercase{Nahian Ibn Asad}\authorrefmark{1}\authorrefmark{2}\authorrefmark{*},
    \uppercase{Md. Mushfiqul Haque Omi}\authorrefmark{1}\authorrefmark{3}\authorrefmark{*},
    \uppercase{Md. Bakhtiar Hasan}\authorrefmark{1},
    \uppercase{Sabbir Ahmed}\authorrefmark{1}\IEEEmembership{Member, IEEE}, 
    and 
    \uppercase{Md. Hasanul Kabir}\authorrefmark{1}\IEEEmembership{Member, IEEE}
}
\address[1]{Department of Computer Science and Engineering, Islamic University of Technology, Dhaka, Bangladesh}
\address[2]{TigerIT Bangladesh Ltd., Dhaka, Bangladesh}
\address[3]{Department of Computer Science and Engineering, United International University, Dhaka, Bangladesh}
\address[*]{These authors have contributed equally to this work.}
\corresp{Corresponding author: Md. Hasanul Kabir (e-mail: \hyperlink{mailto:hasanul@iut-dhaka.edu}{hasanul@iut-dhaka.edu}).}
\tfootnote{This work was supported in part by the IUT Seed Grant 2022 under grant no. REASP/IUT-RSG/2022/OL/07/011.}

\markboth
{Rahman \headeretal: \modelName: Detecting Traffic Signs with Limited Data}
{Rahman \headeretal: \modelName: Detecting Traffic Signs with Limited Data}

\begin{abstract}
Automatic Traffic Sign Recognition is paramount in modern transportation systems, motivating several research endeavors to focus on performance improvement by utilizing large-scale datasets.
As the appearance of traffic signs varies across countries, curating large-scale datasets is often impractical; and requires efficient models that can produce satisfactory performance using limited data.
In this connection, we present `\modelName', built-upon \textbf{F}aster RCNN for traffic sign detection, enhanced by \textbf{U}nfrozen Parameters, Pseudo-\textbf{S}upport Sets, \textbf{E}mbedding Normalization, and \textbf{D}omain Adaptation while reducing data requirement. 
Unlike traditional approaches, we keep all parameters unfrozen during training, enabling \modelName\ to learn from limited samples. The generation of a Pseudo-Support Set through data augmentation further enhances performance by compensating for the scarcity of target domain data.
Additionally, Embedding Normalization is incorporated to reduce intra-class variance, standardizing feature representation. 
Domain Adaptation, achieved by pre-training on a diverse traffic sign dataset distinct from the target domain, improves model generalization.
Evaluating \modelName\ on the BDTSD dataset, we achieved $2.4\times$, $2.2\times$, $1.5\times$, and $1.3\times$ improvements of mAP in 1-shot, 3-shot, 5-shot, and 10-shot scenarios, respectively compared to the state-of-the-art Few-Shot Object Detection (FSOD) models. Additionally, we outperform state-of-the-art works on the cross-domain FSOD benchmark under several scenarios.
\end{abstract}


\titlepgskip=-21pt

\maketitle
\section{Introduction}
Traffic sign recognition plays a crucial role in enhancing road safety, supporting autonomous vehicle technology, and assisting in efficient driving \cite{Gudigar2016aReview}.
However, interpreting traffic signs in real-time can be challenging because of issues such as environmental factors, distance, and limited visibility \cite{liu2019machine}, which can lead to misrecognition and potentially catastrophic accidents \cite{mehran2022understanding}. To mitigate these risks, Advanced Driver Assistance Systems \cite{paul2016advanced} and Automated Driving Systems \cite{chan2017advancements} have been developed in the literature, with traffic sign detection being a key component of their functionality. Hence, accurate traffic sign detection is vital for the effective operation of these systems, especially in real-world conditions.

Over the years, traffic sign detection systems have evolved from manually engineered features to Deep Learning (DL)-based methods, particularly Convolutional Neural Networks (CNNs) \cite{Kheder2024improved}. These methods are generally categorized into two categories: one- and two-stage detectors \cite{kang2022survey}. One-stage detectors such as YOLO \cite{redmon2016you, verghese2024yolov8} and DETR \cite{carion2020end} treat sign detection as a regression problem. Here, a single convolutional network simultaneously predicts multiple bounding boxes around the traffic sign along with the class probabilities for those boxes, offering faster processing but less accuracy \cite{alamgir2022performance, rahman2022densely}. In contrast, two-stage detectors, such as Faster R-CNN \cite{ren2015faster}, first generate region proposals around the signs and then classify them, leading to higher accuracy but slower inference.

Despite the emergence of DL-based methods that have introduced significant progress in traffic sign detection \cite{zhu2016traffic, tabernik2019deep}, these methods require extensive annotated training data to achieve optimal performance.
However, publicly available datasets, such as TT100K \cite{zhu2016traffic}, primarily represents traffic signs from specific regions like China, which are significantly different from those in underdeveloped or developing countries, such as Bangladesh. In these countries, the condition of traffic signs is often compromised due to limited maintenance, occlusion, or wear and tear, introducing additional challenges that are not well-represented in such datasets.
Consequently, a model trained on one country's dataset might not perform well on another without being fine-tuned on a sufficient number of target samples. In this context, collecting large-scale annotated datasets for traffic signs across every country is impractical due to the diversity of sign designs and environmental conditions.

Even with the advancements in hardware technology and the generation of massive datasets that have propelled the DL-based approaches towards human-like capabilities, they are still unable to learn effectively from only a few samples \cite{qi2022smallData}. Unlike humans, who can quickly adapt to new tasks using prior knowledge, DL models require extensive data to perform effectively \cite{Antonelli2022fewshot}. This disparity in learning efficiency is particularly challenging in the context of traffic sign detection in underrepresented regions, such as Bangladesh, where annotated samples are scarce. In this connection, the recent family of Few-Shot Learning (FSL)-based models can play an instrumental role with its remarkable ability to learn with limited data and has already proved its worth in a diversified arena of computer vision tasks \cite{song2023aComprehensive}, making them an ideal choice for this problem domain.

Recent research endeavors have applied FSL-based approaches to problems such as object detection and classification, collectively referred to as Few-Shot Object Detection (FSOD) \cite{liu2023recent}. FSOD is especially useful in situations where it is difficult to obtain large labeled datasets, such as with traffic signs from remote locations that are underrepresented due to negligence and lack of preservation efforts \cite{ashik2022recognizing}. Hence, in this work, we propose \modelName, a fusion of Faster RCNN with Unfrozen Parameters, Pseudo-Support Sets, Embedding Normalization, and Domain Adaptation while reducing data requirement. 
Empirical analysis shows that our proposed model \modelName\ achieves superior performance compared to the state-of-the-art FSOD architecture on the target traffic sign dataset with limited data, significantly reducing the need for large datasets.
This advancement can greatly improve Traffic Sign Detection and Recognition Systems, allowing them to identify traffic signs accurately in various real-world conditions, tackling the challenges of data scarcity. Our contributions are:

\begin{itemize}
    \item Unlike conventional few-shot object detection models that selectively freeze parameters during training, we propose an architecture where the entire network remains unfrozen throughout the training process. This approach allows the model to fully adapt to the limited number of available samples, enabling more robust learning and representation extraction from the few examples in the target domain.
    \item To mitigate the challenge of data scarcity inherent in few-shot learning, we leverage a novel technique called pseudo-support sets that are generated by applying data augmentation techniques to the few available labeled samples per class. This approach not only increases the diversity of the training data but also significantly enhances the model's detection performance by compensating for the limited sample size in the target domain.
    \item We incorporate a cosine similarity-based classifier in our architecture to implement Embedding Normalization (EN). By reducing intra-class variance, EN ensures that the feature representations of traffic signs are more standardized and consistent across the limited number of samples. This improves the model's ability to differentiate between subtle class differences, leading to more precise detection and classification.
    \item We enhance the generalization capability of the proposed architecture through a Domain Adaptation strategy. By pre-training the model on a diverse traffic sign dataset that is distinct from the target query set, we enable the model to better adapt to the specific characteristics of the target domain. This approach significantly boosts performance on datasets with minimal overlap between the training and test domains, proving crucial for cross-domain few-shot detection tasks.
\end{itemize}

The remainder of this paper is organized as follows. In \autoref{sec:relatedWorks}, we review the relevant literature, focusing on existing approaches to Few-Shot Traffic Sign Detection and Few-Shot Object Detection in general, highlighting the current challenges and limitations. \autoref{sec:proposedMethodology} presents our proposed \modelName, detailing the baseline and the modifications made to the Faster R-CNN framework. In \autoref{sec:resultsAndDiscussion}, we conduct extensive experiments on multiple datasets demonstrating the performance of \modelName\ in various few-shot scenarios and compare the results with the state-of-the-art FSOD models. We also discuss the effectiveness of our key contributions and cross-domain generalization performance. \autoref{sec:conclusion} concludes the paper, summarizing our findings and offering directions for future research in few-shot traffic sign detection.

\section{Related Works}\label{sec:relatedWorks}
Few-Shot Object Detection (FSOD) has emerged as a critical research area due to the growing need for models that can effectively detect objects with minimal labeled data \cite{Antonelli2022fewshot}. As traditional object detection methods typically require large-scale annotated datasets to achieve high performance, FSOD seeks to overcome this limitation by enabling models to generalize well even when only a few instances of the target objects are available. 

The literature on FSOD is extensive and diverse, with several approaches being proposed and refined over the years. According to \cite{kohler2023few}, FSOD techniques can be broadly categorized into three primary categories: transfer learning-based, meta-learning-based, and metric learning-based approaches.
Transfer learning in FSOD leverages a pre-trained network's feature extraction capabilities, which are fine-tuned on a smaller target dataset to adapt its features to a new domain with minimal additional data \cite{wang2020frustratingly}. This method is powerful because it applies rich features learned from large datasets to specific tasks with limited data. However, ensuring effective generalization to new classes without overfitting remains a challenge. Regardless, simple fine-tuning has sometimes outperformed complex methods like meta-learning, indicating their robustness. 

Meta-learning, commonly known as, ``learning to learn'', enables rapid adaptation to new detection tasks by training models on diverse range of object classes. However, as highlighted by \cite{wang2020frustratingly}, these methods often require complex architectures and extensive training. At the same time, their effectiveness can be limited by inconsistent evaluation standards in the field, making them less practical for real-world FSOD applications. Metric learning, focused on learning a distance function to differentiate object classes, excels in FSOD by allowing efficient comparison of new instances with few labeled examples. Regardless of their simplicity and efficiency, a fundamental issue with metric learning is that they tend to perform poorly when there are slight visible changes in the support set or query set \cite{li2023deep}.

Despite the progress in FSOD, there remains a notable gap in the literature regarding few-shot traffic sign detection. Most research in this domain has focused on traffic sign classification \cite{zhou2020few,nguyen2024self}, with very few studies addressing the unique challenges of traffic sign detection. Notable exceptions include \cite{ren2022meta} and \cite{sun2024transformer}, who explored FSOD in traffic scenarios using YOLO-based and transformer-based architectures, respectively. However, these approaches have generally underperformed compared to Faster R-CNN-based architectures, as discussed by \cite{kohler2023few} and \cite{xin2024few}. YOLO-based models, while faster, often struggle with localization accuracy, which is critical in detecting small and closely spaced objects like traffic signs and the performance is even more degraded when the data is scarce. Transformers, on the other hand, incur a high computational cost, making them less practical for real-time detection tasks in traffic environments and they also pose the requirement of abundant data for high performance. These limitations have led us to adopt Faster R-CNN as our base detector, given its superior performance in terms of localization and its ability to be effectively adapted for few-shot learning through transfer and metric learning techniques \cite{ren2015faster, fan2020few, wang2020frustratingly, qiao2021defrcn, xiong2023cd}.

\begin{figure}[tb]
\centering
\includegraphics[width=\textwidth]{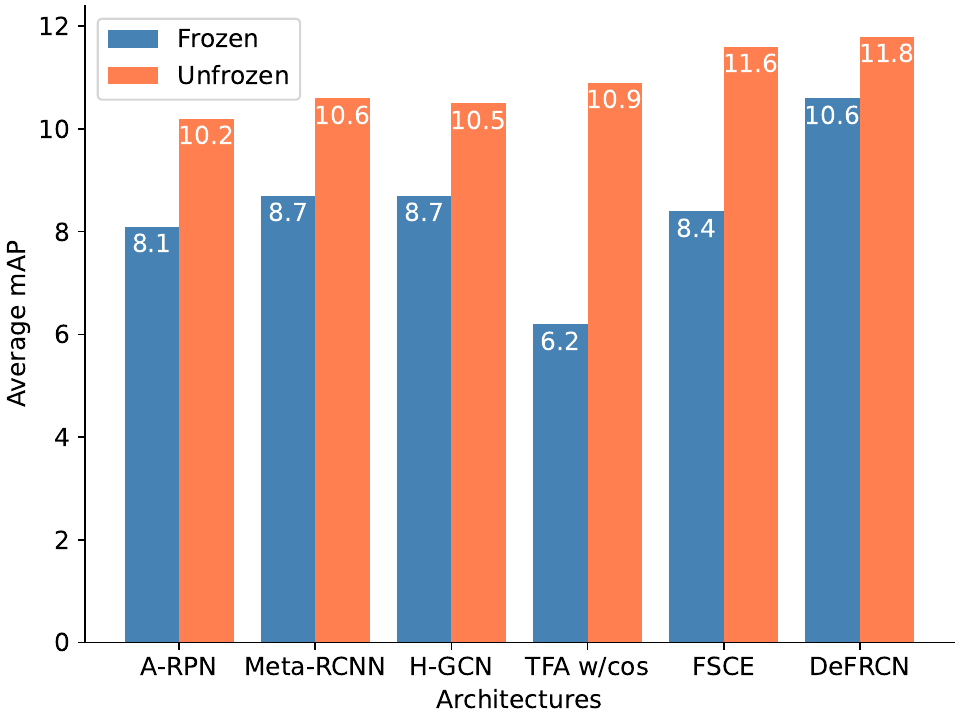}
\caption{Comparison of Mean Average Precision (mAP) between frozen and unfrozen training conditions across different state-of-the-art architectures on the Cross-domain FSOD benchmark. The results demonstrate that keeping parameters unfrozen during training consistently enhances performance, as indicated by the higher mAP scores in the unfrozen condition. (Data adapted from \cite{xiong2023cd})}
\label{fig:impactOfUnfrozenParameters}
\end{figure}

Among the approaches built upon Faster R-CNN, TFA w/cos \cite{wang2020frustratingly} represents a significant advancement in FSOD by introducing a two-stage detector with a frozen backbone and a cosine similarity-based classifier. This approach was designed to prevent overfitting to the small target dataset by keeping most of the network's parameters fixed during fine-tuning. However, subsequent studies, such as FSCE \cite{sun2021fsce}, have highlighted the limitations of this approach, particularly the potential performance degradation caused by freezing key modules like the Region Proposal Network (RPN). FSCE addressed these issues by introducing a Contrastive Proposal Encoding Loss (CPE Loss) that enhances the model's ability to differentiate between similar and dissimilar proposals, thus improving intra-class compactness and inter-class variance. However, this architecture still faced challenges in generalizing to novel classes, which is a critical aspect of FSOD.

The decision to freeze certain modules in FSOD architectures has been a subject of debate in the literature. Cross-Domain Few-Shot Object Detection (CD-FSOD) \cite{xiong2023cd}, for instance, argues that unfrozen parameters are crucial for enhancing the performance of FSOD models, particularly in cross-domain scenarios. The authors showed that architectures with unfrozen parameters consistently outperformed those with frozen modules, suggesting that the flexibility to update the parameters is key to adapting to new and diverse object categories. As shown in \autoref{fig:impactOfUnfrozenParameters}, unfrozen parameters can significantly enhance performance, particularly in the context of few-shot traffic sign detection, where the diversity of object instances is relatively limited compared to broader object detection tasks.

One key takeaway from TFA w/cos approach is the utilization of a cosine similarity-based classifier to normalize and compare instance-level embeddings, enhancing the model's ability to capture relationships between instances and classes. As recommended by \cite{wang2020frustratingly}, this approach ensures comparability across different instances, improving generalization and stability in dynamic few-shot learning scenarios where the model must adapt to new classes without forgetting previous ones \cite{luo2018cosine}. This led us to incorporate a cosine-similarity-based classifier in our work to improve performance in data-scarce environments by leveraging this effective embedding normalization.

Again, as per the observations of \cite{xiong2023cd, fan2020few}, domain adaptation and data augmentation are crucial strategies for improving the generalization of FSOD models, particularly in cross-domain settings where the target categories differ significantly from those in the training data. The CD-FSOD benchmark exemplifies this approach by incorporating a diverse set of categories from different domains, thereby challenging the models to adapt to a broader range of object classes. 
The CD-FSOD model also employed data augmentation techniques, including both strong and weak augmentations, to enhance the robustness of their models. These techniques, such as color jittering, grayscale conversion, gaussian blur, and cutout patches, help prevent overfitting by introducing variability into the training data. However, the complexity of the CD-FSOD approach, particularly the use of a teacher-student distillation framework and Exponential Moving Average (EMA) updates, may be unnecessary for tasks like traffic sign detection, where the objects have relatively consistent shapes and colors. 
Our analysis showed that a simpler FSOD model, based on a straightforward Faster R-CNN fine-tuning (FRCN-ft) approach, retained a significant portion of CD-FSOD's performance while being easier to implement and more computationally efficient. This inspired us to adopt a simpler methodology that utilizes an augmented support set, named pseudo-support set to improve detection performance in our target domain.

The use of attention mechanisms in FSOD has also gained traction, particularly in the work by \cite{fan2020few}, who proposed an Attention-based Region Proposal Network (A-RPN) utilizing information from the support set images that enables them to focus on the relevant portion of the query set images during inference. This approach aims to improve the model's ability to detect novel categories by directing attention to relevant regions in the query images. However, our evaluation of A-RPN revealed its poor performance in few-shot traffic sign detection settings. Upon further investigation, we identified a significant disparity in the quality of support images generated by the attention RPN when fine-tuning with the Bangladeshi Traffic Sign Detection Dataset (BDTSD) \cite{ashik2022recognizing}. This indicates that the effectiveness of attention mechanisms in FSOD highly depends on the quality of the support images, which may not always be guaranteed in real-world scenarios. Consequently, while attention mechanisms hold promise, we chose not to incorporate them into \modelName\ due to the performance inconsistency observed in our evaluations.

Decoupled Faster R-CNN (DeFRCN) \cite{qiao2021defrcn} represents another significant contribution to the FSOD field by improving the standard Faster R-CNN architecture to make it more suitable for few-shot learning tasks. DeFRCN introduced multi-stage decoupling and enhanced translation invariant features via Gradient Decoupled Layer (GDL) and Proto Calibration Block (PCB), respectively. Despite these advancements, DeFRCN's reliance on frozen modules and its inability to enhance classification scores with low-quality region proposals limit its applicability in scenarios where high-quality proposals are not guaranteed. This underscores the importance of robust proposal generation in FSOD, particularly for tasks like traffic sign detection, where the quality of the region proposals can significantly impact the model's performance.

In conclusion, the extensive body of research on FSOD highlights the diversity of approaches and the ongoing debates regarding the most effective strategies for few-shot learning. Despite the progress made so far, significant challenges remain, particularly in the application of FSOD to specific domains like traffic sign detection. Drawing inspiration from the strengths and limitations of the existing literature, we propose \modelName, which integrates effective strategies from state-of-the-art architectures while introducing novel enhancements. Our methodology leverages the robustness of Faster R-CNN, the adaptability of transfer and metric learning, and the practical benefits of domain adaptation and pseudo-support sets, all tailored to address the unique challenges of few-shot traffic sign detection. The following sections provide a detailed exploration of our proposed approach, demonstrating how \modelName\ advances the state-of-the-art in this specialized field.

\section{Proposed Methodology}\label{sec:proposedMethodology}

\begin{figure*}[t]
\centering
\includegraphics[width=0.94\textwidth]{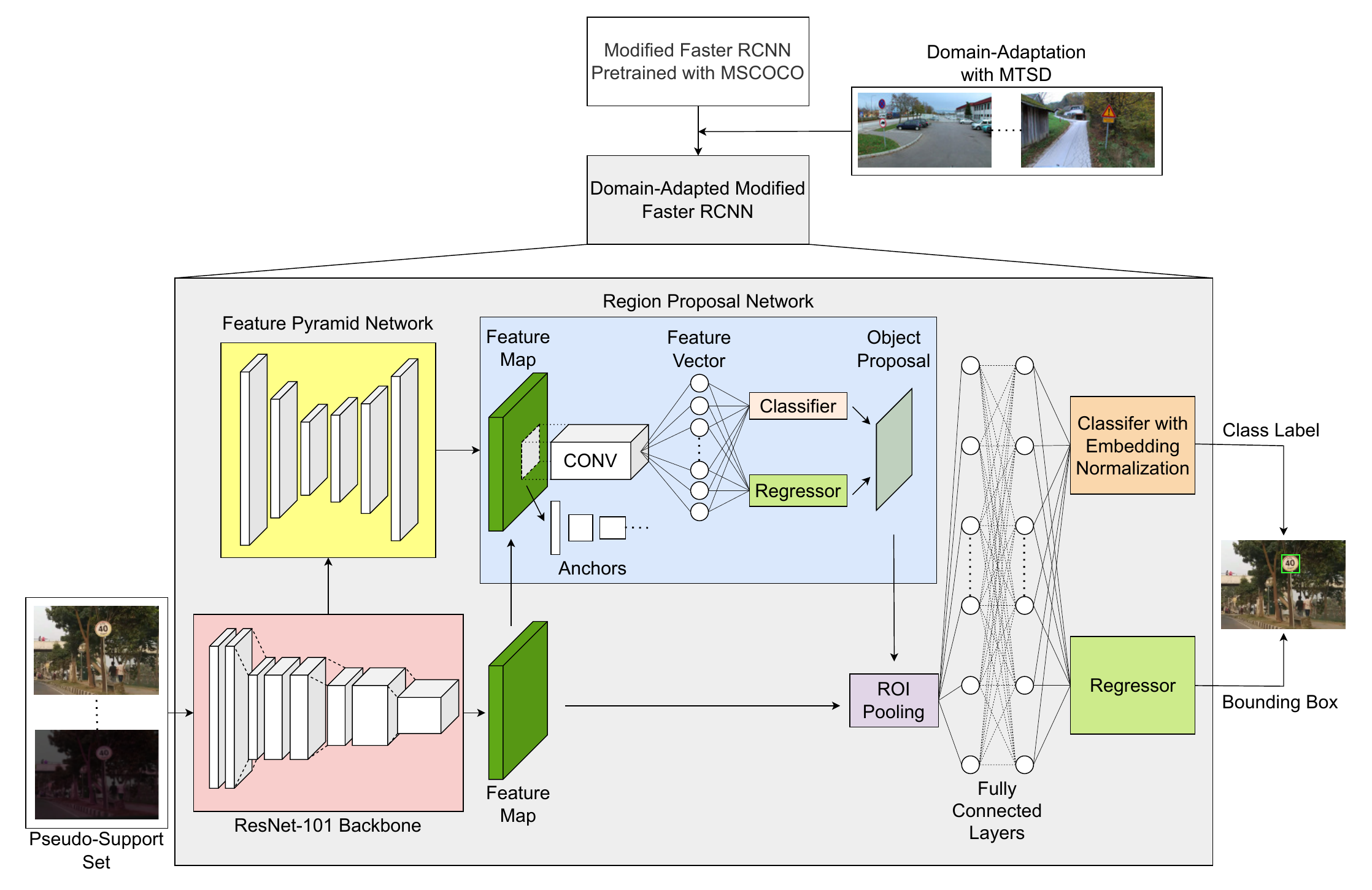}
\caption{Overview of the \modelName\ Architecture. The architecture begins with a Faster R-CNN model pre-trained on the MSCOCO dataset, which is then modified by incorporating a cosine similarity-based classifier for embedding normalization. This modified Faster R-CNN undergoes domain adaptation using the Merged Traffic Sign Detection Dataset (MTSD), enhancing its ability to handle diverse traffic sign appearances. In the fine-tuning stage, the model learns from a pseudo-support set generated from the target Bangladeshi Traffic Sign Detection Dataset (BDTSD). Throughout the process, all modules of the model remain unfrozen.}
\label{fig:pipeline}
\end{figure*}

Our proposed framework is built upon Faster R-CNN, pre-trained on the MS COCO dataset \cite{lin2014microsoft}, due to its established robustness in object detection tasks \cite{desai2020survey}. Recognizing the challenges inherent in few-shot traffic sign detection, we introduced several key modifications to enhance the model's ability to generalize from a small number of examples. First, we keep the entire network unfrozen during training, allowing all layers to adapt to the new data and enabling more effective learning of domain-specific features. Second, we augment the training data by creating a pseudo-support set, which combines original samples with their augmented versions, increasing diversity and improving the model's resilience to real-world variations. To further mitigate the high intra-class variance inherent in few-shot scenarios, we replace the original classifier with a cosine similarity-based classifier. This replacement incorporates embedding normalization, focusing on the intrinsic relationships between instances and classes to ensure better generalization from sparse training samples.

Additionally, we fine-tune the entire network end-to-end using the MTSD dataset, ensuring that the features learned are specifically tailored to traffic signs, which present unique characteristics distinct from the general objects in COCO. The effectiveness of these strategies is evaluated by testing FUSED-Net on the target dataset across various few-shot scenarios (1, 3, 5, and 10-shot), demonstrating its suitability for the specialized task of traffic sign detection in data-scarce environments. The overview of the FUSED-Net Architecture is shown in \autoref{fig:pipeline}.


\subsection{Baseline Faster-RCNN}
To thoroughly understand the strategies we leveraged to adapt Faster R-CNN for few-shot traffic sign detection, it is crucial to first grasp the architecture of Faster R-CNN and its evolution from earlier models. The foundation of Faster R-CNN lies in the progression from RCNN (Region-based Convolutional Neural Network) \cite{girshick2014rich} to Fast R-CNN \cite{girshick2015fast}, each of which introduced key innovations in object detection.



The object detection process traditionally involves three steps: generating region proposals, extracting feature vectors, and classifying them. Early models like RCNN used Selective Search \cite{uijlings2013selective} for proposal generation and CNNs for feature extraction, improving accuracy over traditional descriptors. However, RCNN relied on a pre-trained Support Vector Machine \cite{cortes1995support} for classification, which was computationally expensive due to the need for caching features from each region proposal. Fast R-CNN improved upon this by introducing the ROI pooling layer, enabling the extraction of fixed-size feature vectors from a shared feature map, allowing proposals to share computation. This innovation sped up the process and eliminated the need for caching, significantly streamlining the detection pipeline.

Building on Fast R-CNN, Faster R-CNN \cite{ren2015faster} introduced the Region Proposal Network (RPN), a fully convolutional network, designed to predict region proposals directly from the feature maps, thus integrating the proposal generation process into the neural network. This eliminated the need for external algorithms like selective search, making the detection process both faster and more accurate. Additionally, Faster R-CNN introduced a pyramid of reference anchor boxes, which allowed it to handle objects of varying scales and aspect ratios within a single framework. These multi-scale anchors enabled the sharing of features between the RPN and the classifier, further optimizing computational efficiency.

In modern implementations, Faster R-CNN typically employs deep neural networks like ResNet \cite{he2016deep} or VGG \cite{liu2015very} as the backbone for feature extraction, often enhanced with a Feature Pyramid Network (FPN). These backbones are crucial for generating robust, multi-scale feature representations, which are particularly important in detecting objects that vary in size and shape. The integration of the FPN allows the network to exploit features at multiple scales, improving detection accuracy for small and large objects alike.

Given the widespread success of Faster R-CNN in various object detection tasks \cite{maity2021Faster,Alanezi2022livestock, rahman2022twoDecades, bashar2022multipleobjecttrackingrecent}, many FSOD frameworks \cite{wang2020frustratingly, fan2020few, qiao2021defrcn, xiong2023cd, ren2015faster}, including our own, are built upon this architecture. Specifically, our approach uses a variant of Faster R-CNN provided by the Detectron2 Framework\footnote{\href{https://detectron2.readthedocs.io/en/latest/}{https://detectron2.readthedocs.io/en/latest/}}, with ResNet-101 and FPN as the backbone. This combination was chosen for its ability to produce deep, rich feature representations and its effectiveness in multi-scale feature extraction.

\subsection{Unfrozen Parameters}


In deep learning, ``freezing'' certain layers during training is common, especially when fine-tuning pre-trained models on a new task \cite{peters2019tunetune}. Freezing keeps the weights of those layers fixed, updating only the unfrozen layers. This helps retain learned features while minimizing computational overhead, particularly when the new task is similar to the original task \cite{girshick2014rich, yosinski204how, razavian2014CNN}. However, in cross-domain tasks like few-shot traffic sign detection, where the target domain (e.g., region-specific traffic signs) differs significantly from the base domain (e.g., general object detection), freezing layers can hinder performance \cite{he2016deep, chen2019closer, long2015learning}. The model may struggle to learn new features essential to the target domain, as frozen layers may carry biases and features from the base domain, reducing detection accuracy \cite{liu2018rethinking}.

To overcome this limitation, we chose to keep the entire architecture unfrozen during the fine-tuning stage. This means that all components of our detector, including the Backbone, Region Proposal Network (RPN), Feature Pyramid Network (FPN), and fully connected layers, are updated during training. Given the network with layers $L_1, L_2, \dots, L_n$, unfrozen training updates the weights of all the layers such that $\nabla W_{L_i} \neq 0$ for $i = 1, \dots, n$. By allowing every part of the network to adapt, we ensure that the model can learn new, domain-specific features that are essential for detecting traffic signs in different environments. This approach contrasts with the conventional method of freezing certain layers and allows our model to better handle the significant differences between the base and target domains. As a result, our model demonstrates improved adaptability and accuracy in the target domain.

\subsection{Pseudo Support Sets (PSS)}
In FSOD, the limited availability of labeled training data presents a significant challenge in developing robust models. To address this, data augmentation is a common technique used to enhance the diversity and variability of the training data. In our methodology, we implemented a data augmentation strategy during the fine-tuning stage by generating a `Pseudo-Support Set (PSS)' by augmenting the original support set. Mathematically, let $X_s$ represent the original support set and $T$ represent the transformation applied. The augmented support set $X_s'$ is then defined as:
$$X_s' = X_s | T(X_s)$$
The creation of this pseudo-support set involves applying random Color Jitter to each sample in the original support set. Here, we chose to use only Color Jitter for augmentation, guided by \cite{chen2020simple}, who showed that stronger color jittering is effective in tasks requiring visual representations.

Color Jitter is a data augmentation technique that randomly alters the brightness, contrast, saturation, and hue of an image. These specific transformations were chosen because they closely mimic the kinds of variations that are commonly encountered in real-world scenarios, such as changes in lighting conditions, different times of the day, weather conditions, and even camera settings. Brightness adjustment simulates conditions ranging from very bright, sunny environments to dim, overcast scenarios. Contrast alteration affects the differentiation between light and dark areas in an image, allowing the model to recognize objects in both high-contrast situations like strong shadows, and low-contrast environments like foggy weather. Saturation changes the intensity of colors, helping the model learn to detect traffic signs that might appear more muted or vividly colored due to different camera sensors or environmental conditions. Hue adjustment shifts the overall color balance of the image, which can account for the color variances caused by different lighting sources (e.g., fluorescent vs. natural light).



The creation of PSS by supplementing original samples with their augmented versions introduces a broader range of visual characteristics into the training data. This enriched dataset helps the model recognize patterns under diverse conditions, improving generalization and reducing overfitting. Additionally, training on the PSS enhances the model's robustness, making it less sensitive to input variations and better equipped to handle the unpredictable changes encountered in real-world traffic sign detection, leading to more reliable and accurate performance.

\subsection{Embedding Normalization (EN)} 

Embedding Normalization (EN) is crucial in our method, particularly considering our few-shot learning scenario, where high intra-class variance is a challenge \cite{wang2020frustratingly}. In few-shot scenarios, limited training examples lead to significant variations in feature representations, making the model overly sensitive to outliers and reducing generalization. To address this, EN standardizes feature representations, ensuring that all instances are on a comparable scale, which reduces intra-class variance and enhances robustness.

To incorporate EN, we integrate a cosine similarity-based classifier into FUSED-Net, which focuses on angular relationships between feature vectors, rather than magnitudes. This approach helps mitigate the impact of varying vector lengths, ensuring more accurate detection, especially in few-shot learning where limited data can lead to inconsistent feature magnitudes. EN improves generalization by reducing the influence of outliers and aligning features more effectively, allowing the classifier to distinguish between classes with minimal data. This is particularly beneficial for our application domain, few-shot traffic sign detection, where sparse and diverse data is common.

In our implementation, we normalize each embedding vector $x_i$ such that all instances are on a comparable scale. This standardization facilitates the effective comparison of features across different instances. The cosine similarity between the normalized embedding vector $\hat{x}_i$ and the class weight vector $\hat{w}_j$ is calculated as follows:

$$sim_{(x_i, w_j)} = \gamma \hat{x}_i^T \hat{w}_j$$
where,
\begin{itemize}
    \item $\gamma$ is a scaling factor
    \item $\hat{x}_i$ is the normalized embedding vector of $x_i$
    \item $\hat{w}_j$ is the normalized embedding vector of $w_j$
\end{itemize}

The range of cosine similarity, $sim_{(x_i, w_j)}$ varies between $-1$ to $+1$. A value of $+1$ indicates the vectors $\hat{x}_i$ and $\hat{w}_j$ are perfectly aligned in the same direction, implying a high degree of similarity between the instance and the class. A value of $-1$ suggests that the vectors are diametrically opposed, indicating dissimilarity. A value near $0$ indicates orthogonality, meaning there is no discernible similarity between the two vectors. The value of $\gamma$ is determined empirically.

The normalized embedding vector of $x_i$ is calculated as follows:

$$\hat{x}_i = \frac{x_i}{\|x_i\|}$$
where,
\begin{itemize}
    \item $x_i$ is the embedding vector of the $i^{th}$ instance
    \item $\|x_i\|$ is the magnitude (or length) of the vector $x_i$
\end{itemize}

The normalized embedding vector of $w_j$ is calculated as follows:

$$\hat{w}_j = \frac{w_j}{\|w_j\|}$$
where,
\begin{itemize}
    \item $w_j$ is the weight vector of the $j^{th}$ class
    \item $\|w_j\|$ is the magnitude (or length) of the vector $w_j$\\
\end{itemize}

By considering cosine similarity, we ensure that the model is invariant to vector magnitudes and instead learns to prioritize the orientation of the feature vectors, which is particularly useful for few-shot learning with limited data.

\subsection{Domain Adaptation}
In traditional machine learning, it is generally assumed that the training and testing datasets originate from the same feature space and follow similar joint probability distributions. This assumption is critical for ensuring that the model generalizes well from the training data to unseen test data. However, in many real-world scenarios, this assumption is often violated. For instance, in tasks such as traffic sign detection, the training data may consist of images collected in one country, while the test data could come from a different country with different types of signs, environmental conditions, and even visual styles. This discrepancy in data distribution can lead to a significant drop in the model's performance when applied to new, unseen data.

To address this challenge, the concept of domain adaptation has been introduced in the field of machine learning \cite{farahani2021brief}. Domain adaptation is a subset of transfer learning that specifically aims to reduce the distributional differences between a source domain (training data) and a target domain (test data). The primary objective is to enable the model to generalize effectively to the target domain, despite the differences in feature distributions.

In our approach, we employ domain adaptation by training our model on traffic signs from different countries, thereby exposing it to a diverse set of visual features and styles. This diverse training data helps minimize the distributional discrepancy between the base domain (e.g., traffic signs in the training dataset) and the target domain (e.g., traffic signs from another country with limited data). Let $P_S(X, Y)$ and $P_T(X, Y)$ represent the source and the target distributions, respectively. By training on diverse traffic sign datasets from multiple countries, our model minimizes the discrepancy $\mathcal{D}(P_S, P_T)$. By leveraging this approach, our model becomes more generalized and robust, capable of accurately detecting and recognizing traffic signs across different domains. The effectiveness of this strategy is reflected in the improved performance of our model, which we will discuss in the Results section.

\section{Results \& Discussion}\label{sec:resultsAndDiscussion}
In this section, we present a comprehensive analysis of our proposed FUSED-Net framework. We begin with the experimental setup, which introduces the datasets used, namely Merged Traffic Sign Detection Dataset, Bangladeshi Traffic Sign Detection Dataset, and Cross-Domain Few-Shot Object Detection Benchmark, showcasing the diverse and challenging conditions under which our model was trained and tested. Additionally, we provide the implementation details, including the hyperparameters and experimental environment within the Detectron2 framework. Next, we compare FUSED-Net with the state-of-the-art models in various few-shot traffic sign detection scenarios. This discussion also includes a per-category performance analysis, examining detection accuracy across the traffic sign categories in the query set, and the statistical significance of the reported performance based on multiple independent trials. Following this, we provide a deeper understanding of FUSED-Net's functionality through an ablation study that highlights the incremental benefits of its components and a qualitative analysis illustrating its performance in challenging scenarios using representative examples. Finally, we assess FUSED-Net's adaptability in generalized settings showcasing its results on the Cross-Domain Few-Shot Object Detection (CD-FSOD) benchmark, which underscores the model's robustness and versatility in diverse real-world conditions.

\subsection{Experimental Setup}

\begin{figure}[tb]
    \centering
\subfloat[\label{subfig:euroSpeedLimit}]{
        \includegraphics[width=.8\textwidth]{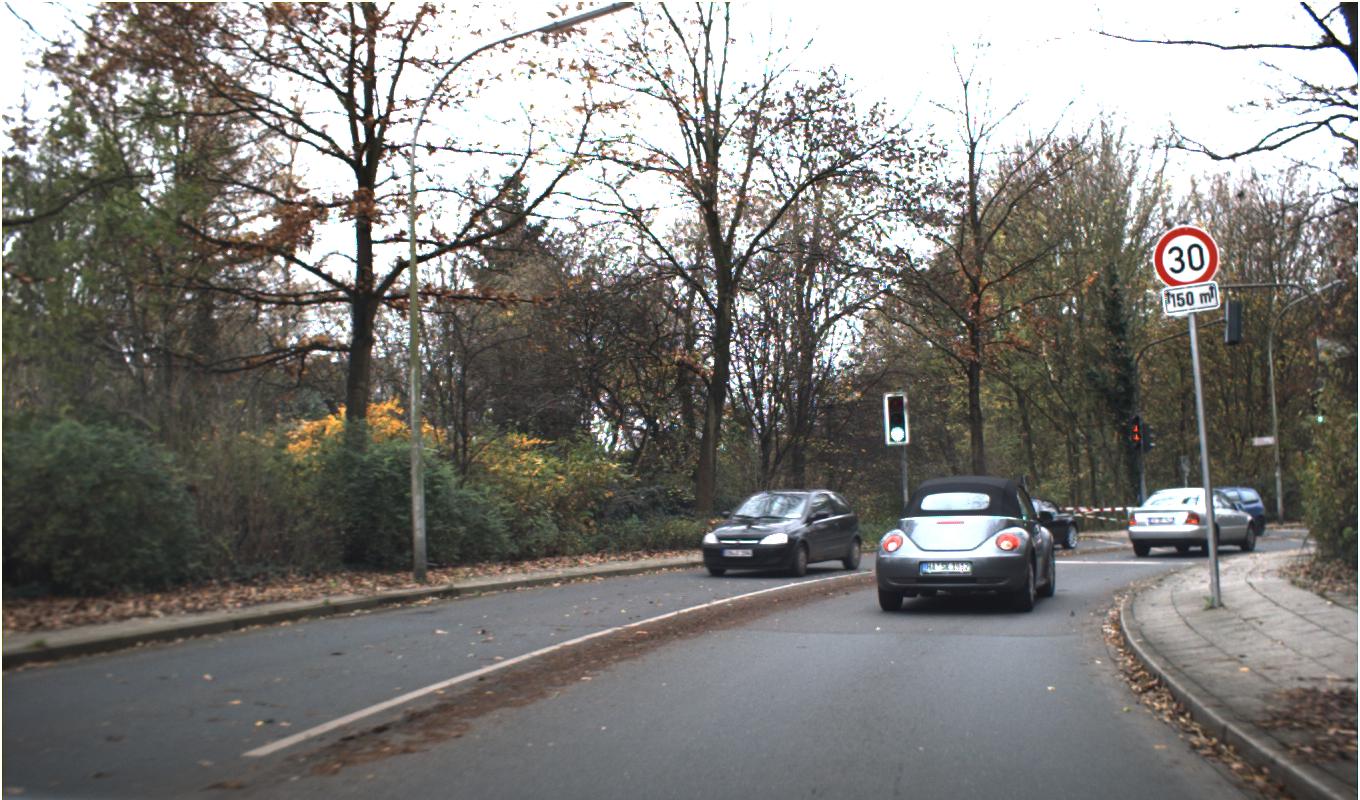}
    }    
    
    \subfloat[\label{subfig:slovaniaLeft}]{
        \includegraphics[width=.8\textwidth]{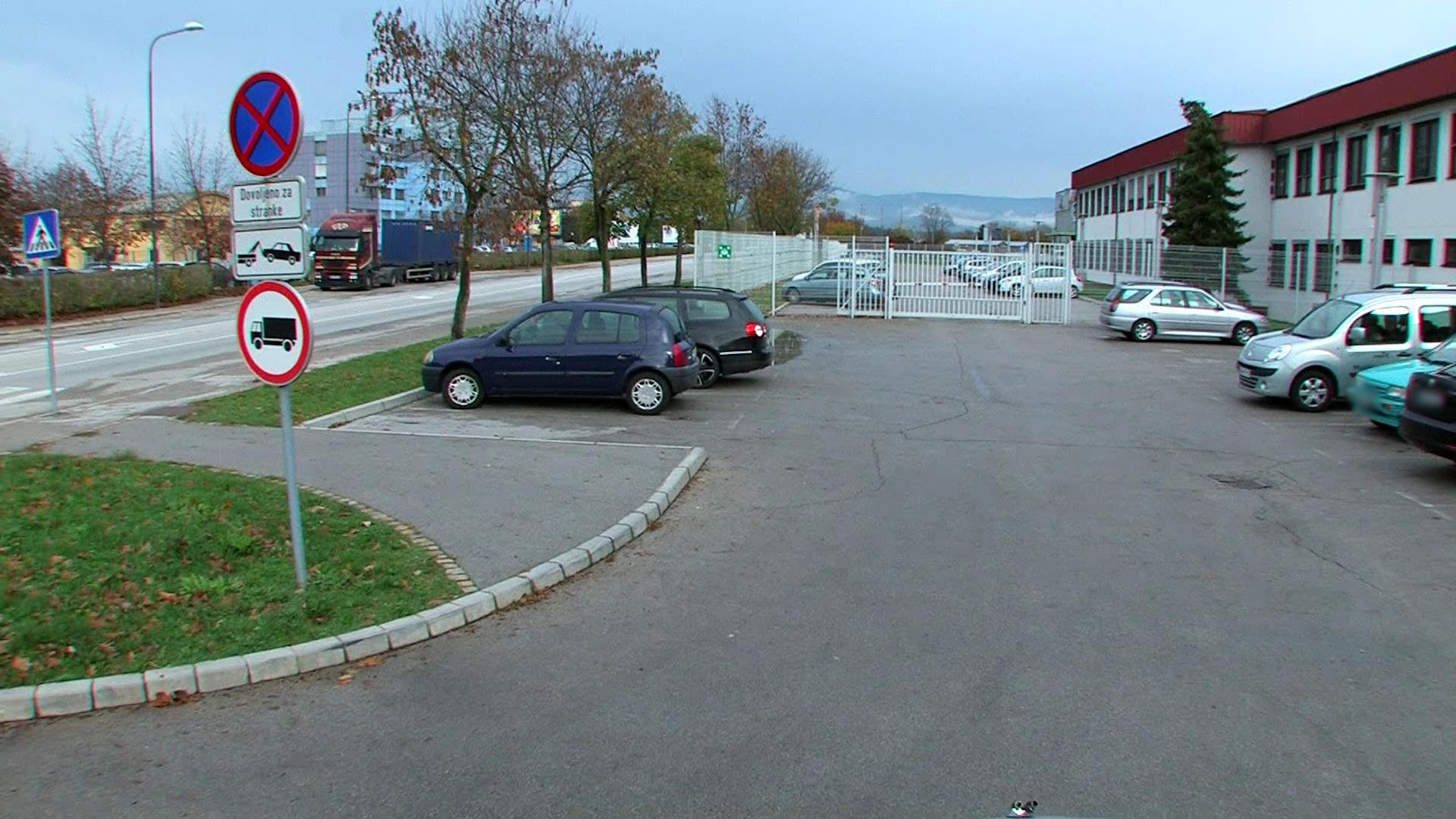}
    }
    
\subfloat[\label{subfig:usSpeedLimit}]{
        \includegraphics[width=.8\textwidth]{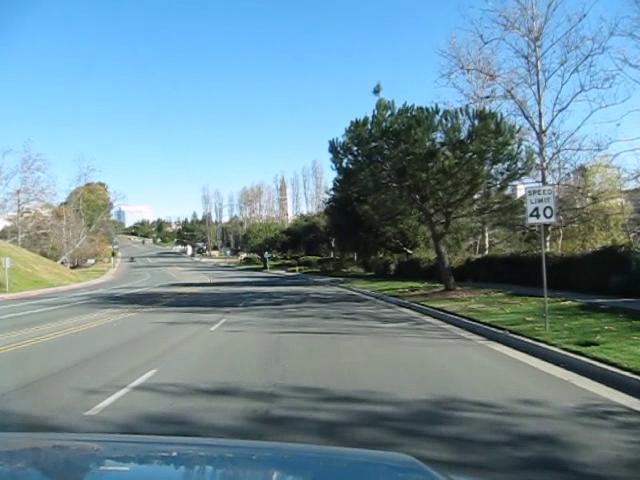}
    }
    \caption{Samples from the Merged Traffic Sign Detection Dataset (MTSD). They illustrate key differences between European (a, b) and U.S. (c) traffic signs. U.S. signs are rectangular with black text on a white background, while European signs are circular with red borders and black numbers. European signs often feature symbols and pictograms, making them more universally accessible, whereas U.S. signs tend to rely on text. Additionally, as evident from (b), European countries may also position traffic signs on the left side of the road and use multiple signs together.}
    \label{fig:mtsdSamples}
\end{figure}

\subsubsection{Datasets}
\paragraph{Merged Traffic Sign Detection Dataset (MTSD)}

The key challenge of FSL models is to generate meaningful feature descriptions with minimal instances, for which it is recommended to be trained on a diverse dataset encompassing a wide range of object categories. The hypothesis is that once the model can describe objects in general, it can provide better feature descriptions even using limited samples of the target domain. 
However, the widely used datasets such as PASCAL VOC \cite{everingham2010pascal} lack traffic sign instances, and MS COCO \cite{lin2014microsoft} includes only a single class for ``stop sign'', making them inadequate for pretraining the baseline models of few-shot traffic sign detection. Drawing inspiration from A-RPN's FSOD dataset \cite{fan2020few}, we constructed the Merged Traffic Sign Detection Dataset (MTSD) by combining samples from three established large-scale traffic sign detection datasets: The German Traffic Sign Detection Benchmark (GTSDB) \cite{houben2013detection}, The LISA Traffic Sign Dataset
(LISA) \cite{mogelmose2012vision}, and DFG Traffic Sign Data Set (DFG) \cite{tabernik2019deep}.

The MTSD dataset is compiled from three established traffic sign detection datasets: The German Traffic Sign Detection Benchmark (GTSDB) \cite{houben2013detection}, the LISA Traffic Sign Dataset (LISA) \cite{mogelmose2012vision}, and the DFG Traffic Sign Dataset (DFG) \cite{tabernik2019deep}. The GTSDB and LISA datasets feature German and American traffic signs, respectively, and together cover a total of 90 categories. Both of these datasets are widely used in the field of traffic sign detection. The DFG dataset, which is unique in its extensive collection, contains 200 categories, making it the largest among the three. In total, the MTSD dataset includes 272 categories, derived from these three sources.

Although MTSD contains 14,416 samples, the presence of multiple traffic signs within a single image leads to a total of 23,276 annotations. To prevent potential confusion in our architecture, we excluded two categories, ``roundabout" and ``stop", which appeared in multiple datasets and had varying sample representations across them. This was done to avoid the possibility of conflicting data affecting model performance.

The dataset was designed with the aim of maximizing sample diversity. To this end, we selected traffic sign datasets from two different continents: Europe (GTSDB, DFG) and America (LISA). Traffic sign designs differ significantly between these regions due to variations in regulatory standards, design philosophies, and cultural norms. For example, U.S. speed limit signs are typically rectangular with black text on a white background (\figureautorefname~\autoref{subfig:usSpeedLimit}), whereas European speed limit signs are circular with a red border and black numbers (\figureautorefname~\autoref{subfig:euroSpeedLimit}). Additionally, European traffic signs often feature symbols and pictograms without text, enhancing cross-linguistic accessibility, while American signs more commonly rely on text, which can be less intuitive. Moreover, in some European countries, traffic signs may be positioned on the left side of the road, and a single location may feature multiple signs (\figureautorefname~\autoref{subfig:slovaniaLeft}).

\begin{figure}[htb]
    \centering
    \subfloat[Number of Samples Per Category]{
        \includegraphics[width=0.85\textwidth]{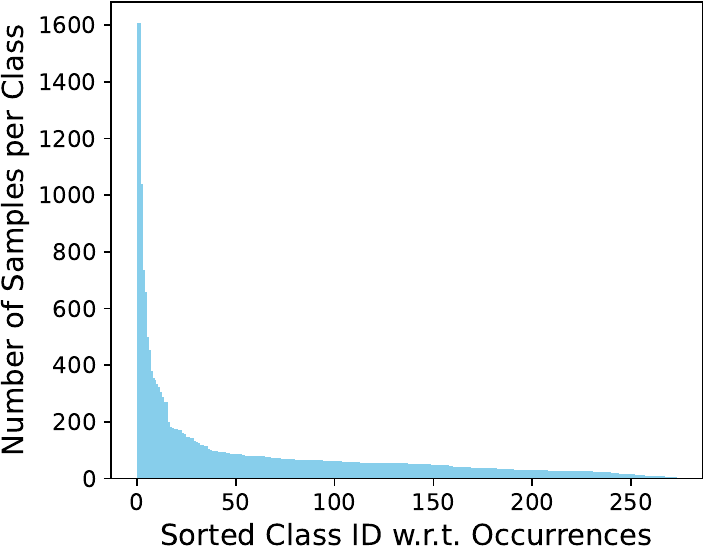}
    }    

    \subfloat[Number of Annotations Per Category]{
        \includegraphics[width=0.85\textwidth]{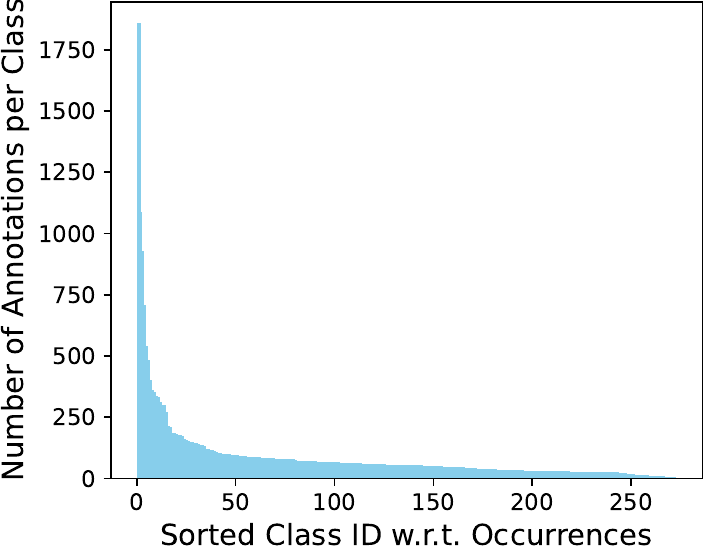}
    }    
    \caption{The dataset statistics of MTSD. The sample and the annotation count per category is highly skewed considering more than 85\% of the categories have less than 100 samples whereas the most frequent categories have more than 1000 samples.}
    \label{fig:mtsdDistribution}
\end{figure}

To provide a clearer understanding of the dataset, we present the class distribution balance in MTSD in \autoref{fig:mtsdDistribution}. The dataset exhibits a highly skewed class distribution: over 85\% of the categories contain fewer than 100 samples, while some categories have more than 1000 samples. This imbalance could impact model performance in few-shot learning tasks, but it does not undermine the dataset's primary purpose. The goal of MTSD is not to offer a balanced training set but to expose the model to a broad range of traffic sign variations from different countries. This diversity is intended to promote the learning of generalized features that can be applied across various visual styles and regulatory standards.

In our proposed framework that incorporates domain adaptation, the MTSD dataset is used for the initial training of the model. This pretraining enables the system to better understand traffic sign characteristics, which enhances its ability to detect unfamiliar signs during the fine-tuning stage. Despite the dataset’s imbalanced nature, this approach helps the model generalize more effectively across domains, improving its performance on real-world traffic sign detection tasks.

\paragraph{Bangladeshi Traffic Sign Detection Dataset (BDTSD)}

\begin{figure*}[tb]
    \centering
    \subfloat[]{
        \includegraphics[width=.45\textwidth]{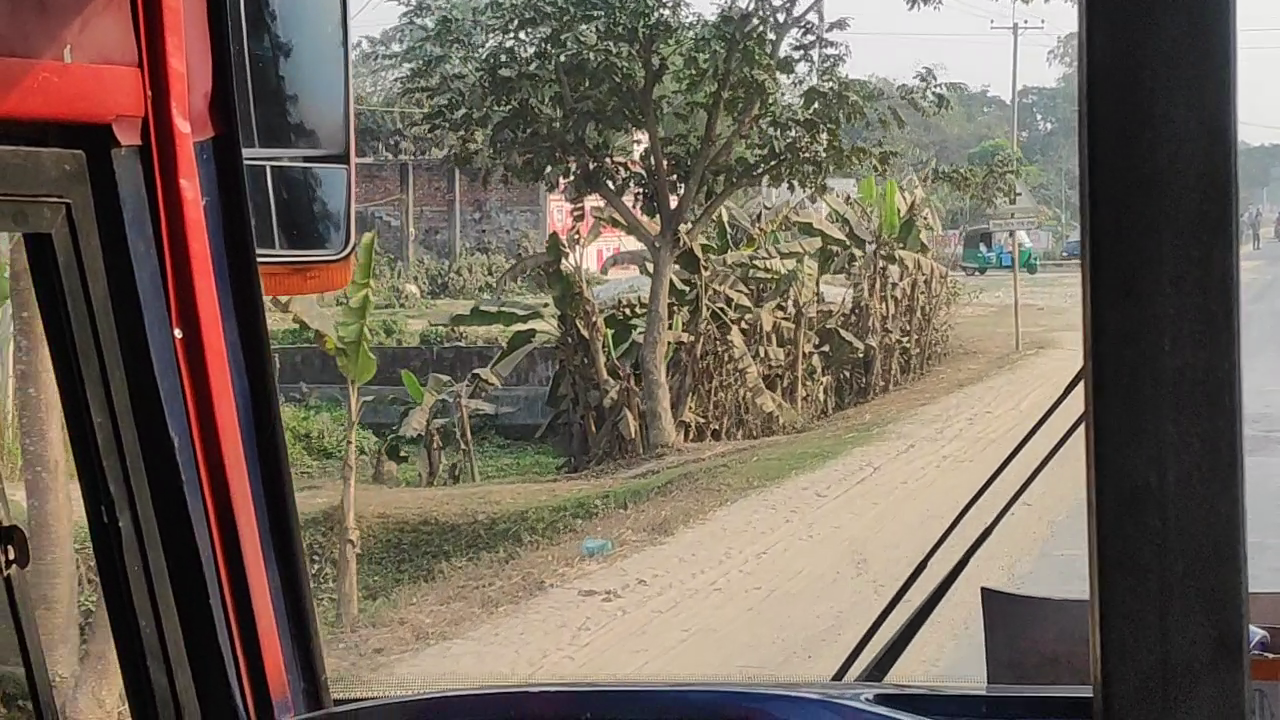}
    }
    \subfloat[]{
        \includegraphics[width=.45\textwidth]{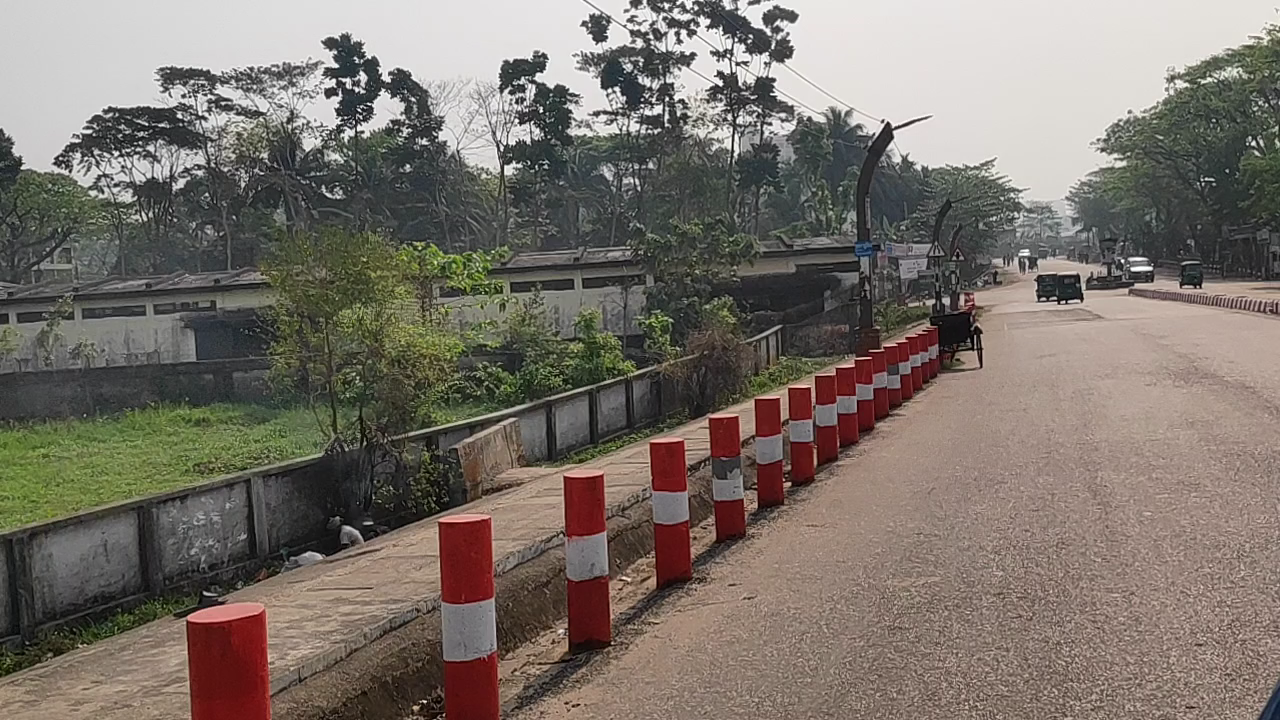}
    }
    
    \subfloat[]{
        \includegraphics[width=.45\textwidth]{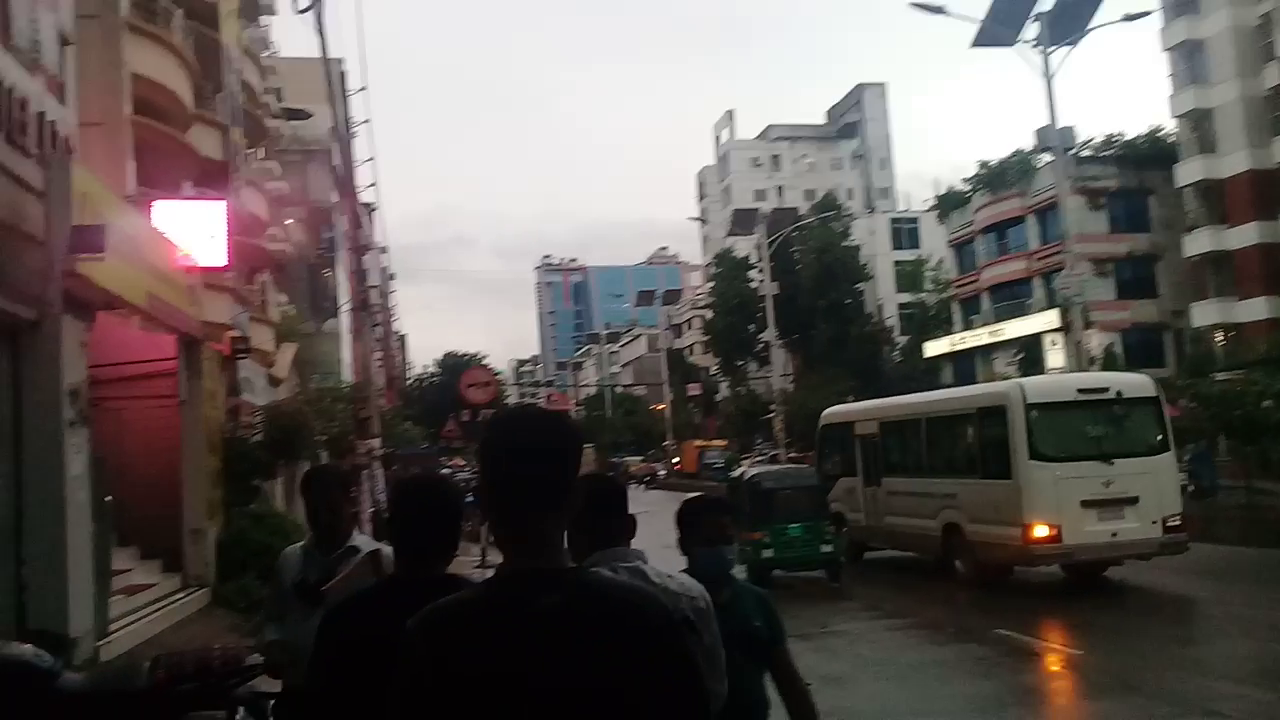}
    }
    \subfloat[]{
        \includegraphics[width=.45\textwidth]{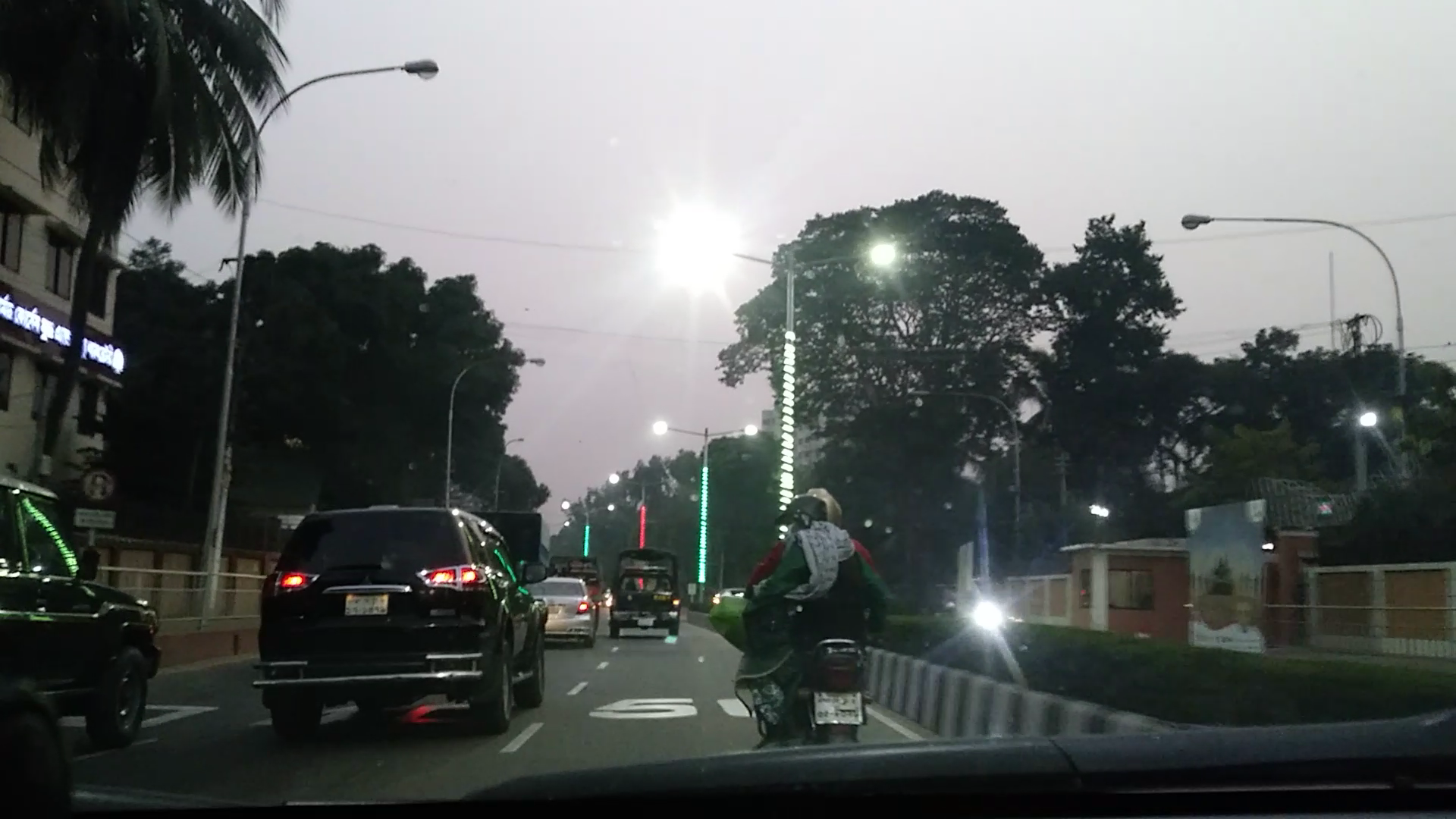}
    }
    \caption{Samples from the Bangladeshi Traffic Sign Detection Dataset (BDTSD). They are characterized by red-bordered signs with numbers and pictograms. The samples highlight various detection challenges. (a) shows a traffic sign partially occluded by trees. The traffic sign in (b) depicts multiple distant traffic signs, complicating detection. (c) presents a blurry traffic sign, while (d) captures a night scene where the sign blends into the background, posing difficulty for detection.}
    \label{fig:bdtsdSamples}
\end{figure*}

The Bangladeshi Traffic Sign Detection Dataset (BDTSD) \cite{ashik2022recognizing} introduces distinct challenges that are not found in standard benchmarks like GTSDB, primarily due to the unique geo-social characteristics of Bangladesh. This dataset comprises 2,986 samples across 15 distinct classes of Bangladeshi traffic signs, collected under a diverse array of real-world conditions. These conditions include variations in distance, occlusion, blur, geological differences, and a range of lighting scenarios, all of which are commonly encountered in practical traffic environments. \autoref{fig:bdtsdSamples} showcases samples that exhibit some of these challenging conditions, providing a visual representation of the complexities involved.

The BDTSD dataset was specifically chosen as the target dataset for both our ablation studies and performance comparisons with other state-of-the-art architectures. This decision was driven by its distinctiveness compared to the datasets used to compile MTSD. While the datasets in MTSD contain more standardized samples, BDTSD reflects the real-world variability and complexity found in Bangladeshi traffic sign scenarios. The aforementioned challenges such as occlusion by other vehicles or objects, signs captured at various distances and angles, and the presence of environmental conditions like rain or fog, can significantly affect the clarity of the signs. Given these unique challenges, BDTSD provides a rigorous testing ground to evaluate the generalization capabilities of our model. By using BDTSD as the target dataset, we ensure that our model is tested against a broad spectrum of conditions, allowing us to assess its robustness and adaptability to real-world situations.

\paragraph{Cross-Domain Few-Shot Object Detection (CD-FSOD) Benchmark}

In few-shot object detection, MS COCO \cite{lin2014microsoft} and PASCAL VOC \cite{everingham2010pascal} are the most commonly used benchmarks for evaluating model generalization. Authors \cite{qiao2021defrcn, wang2020frustratingly} commonly divide PASCAL VOCs 20 categories into 15 bases and 5 novel classes across three random splits, while MS COCO is divided into 60 non-overlapping categories as base classes and 20 as novel classes. However, these benchmarks may not fully reflect real-world scenarios where base and target domain data differ significantly. In practice, few-shot learning aims to address the challenge of limited sample availability from the target domain, a situation not entirely captured by these benchmarks. Recognizing these limitations, the Cross-Domain Few-Shot Object Detection (CD-FSOD) benchmark \cite{xiong2023cd} was proposed as a more realistic and challenging alternative. Unlike traditional benchmarks where both base and novel classes are drawn from the same dataset, the CD-FSOD benchmark constructs its evaluation framework by selecting base and target domain samples from entirely different datasets. This approach ensures that the base and target domains are fundamentally distinct, closely simulating real-world scenarios where the base domain may not adequately represent the target domain. Such a benchmark is critical for testing the true generalization ability of FSOD methods, especially in cases where the target domain is significantly different from the training data.

The CD-FSOD benchmark utilizes MS COCO as its base dataset, leveraging its diverse set of categories and instances. To create a more comprehensive and challenging evaluation, three target datasets from distinct domains are incorporated into the benchmark \cite{xiong2023cd}. The first of these target datasets is Arthropod Taxonomy Orders Object Detection Dataset (ArTaxOr) \cite{drange2019arthropod}, which, while composed of natural samples similar to MS COCO, is specialized in the biological domain. It contains fine-grained categories specific to arthropods, offering 7 classes with 13,991 images for training and 1,383 images for testing. This dataset is particularly valuable for evaluating a model's ability to generalize from broad, diverse categories to fine-grained, domain-specific categories.

The second target dataset in the CD-FSOD benchmark is the Underwater Object Detection Dataset (UODD) \cite{jiang2021underwater}, which introduces challenges from the underwater domain. The samples in UODD are characterized by poor visibility, low color contrast, and a unique visual environment that starkly contrasts with the terrestrial imagery found in MS COCO. UODD consists of 3 classes, with 3,194 images for training and 506 images for testing. This dataset tests the model's robustness to challenging visual conditions and domain shifts that are far removed from typical object detection scenarios.

The third and final target dataset for the CD-FSOD benchmark is Dataset for Object Detection in Aerial Images (DIOR) \cite{li2020object}, which consists of optical remote sensing images. DIOR introduces significant perspective distortion and varying scales, making it the most dissimilar to MS COCO among the three target datasets. It includes 20 classes with 18,463 training samples and 5,000 testing samples. DIOR is particularly challenging due to the overhead perspective, requiring models to adapt to entirely different viewpoints and image structures than those found in the base dataset.

By incorporating such diverse and domain-specific target datasets in our cross-domain testing, we establish a robust platform for evaluating the cross-domain generalization capability of \modelName. The inclusion ensures that the framework is tested under conditions that closely mirror the complexities of real-world applications, where the training and target domains often differ significantly.

\subsubsection{Implementation Details}
\paragraph{Framework and Architecture}
Our proposed framework, FUSED-Net, leverages a variant of Faster R-CNN seamlessly integrated within the Detectron2 framework. Specifically, we employ ResNet-101 in combination with a Feature Pyramid Network (FPN) as the backbone, ensuring robust multi-scale feature extraction. Data augmentation was a critical aspect of our implementation; we applied color jitter with a probability of $0.8$ (as recommended by \cite{chen2020simple}) using the transformations provided by the torchvision library \cite{marcel2010torchvision}, enhancing the diversity and resilience of the training data.

\begin{table*}[tb]
    \centering
    \caption{Comparison of mAP Performance across Different Few-Shot Object Detection Models on the Bangladeshi Traffic Sign Detection Dataset (BDTSD) under 1-Shot, 3-Shot, 5-Shot, and 10-Shot Scenarios. Models are sorted by whether the architecture has RPN or not, and then by the average mAP ($\mu$) across all shots, with the best results highlighted in boldface.}
    \label{tab:comparisonSOTA}
    \begin{tabular}{L{0.16\textwidth} *{5}{R{0.065\textwidth}}}
        \toprule
        \textbf{Model} & \textbf{1-shot} &  \textbf{3-shot} &  \textbf{5-shot} &  \textbf{10-shot} & $\mathbf{\mu}$\\
        \midrule
        \multicolumn{6}{l}{\textit{One-Stage Detectors}}\\
        \midrule
        DETR-ft \cite{carion2020end} & 2.00 & 6.90 & 8.10 & 13.98 & 7.75\\
        YOLOv8-ft \cite{verghese2024yolov8} & 12.80 & 21.80 & 31.20 & 45.50 & 27.83\\
        \midrule
        \multicolumn{6}{l}{\textit{Two-Stage Detectors}}\\
        \midrule
        TFA w/cos \cite{wang2020frustratingly}  & 4.22 & 6.54 & 8.19 & 9.16 & 7.03\\
        A-RPN \cite{fan2020few} & 8.11 & 9.18 & 10.57 & 12.22 & 10.02\\
        DeFRCN \cite{qiao2021defrcn} & 12.53 & 21.93 & 25.84 & 32.28 & 23.15\\
        FRCN-ft \cite{ren2015faster} & 11.36 & 18.58 & 31.19 & 42.12 & 25.81\\
        CD-FSOD \cite{xiong2023cd} & 12.14 & 19.61 & 31.02 & 43.02 & 26.45\\
        \modelName\ (Ours) & \textbf{26.64} & \textbf{44.22} & \textbf{45.94} & \textbf{54.44} & \textbf{42.81}\\
        \bottomrule
    \end{tabular}
\end{table*}

\paragraph{Optimization Strategy}
We utilized the Stochastic Gradient Descent (SGD) solver with a momentum of $0.9$ and a weight decay parameter of $10^{-4}$ to prevent overfitting. During the initial trials, we encountered NAN/INF bounding box errors due to optimization instability caused by the mismatch between the high default learning rate and the low batch size required by our single GPU environment. The small batch size of $2$ introduced significant gradient noise, and the high default learning rate ($0.02$) exacerbated this issue, producing overly aggressive updates that led to unstable bounding box predictions and NAN/INF errors during loss computation.

To address this, we followed Detectron2 recommendations and adjusted the base learning rate and batch size to suit our computational setup. Specifically, we reduced the batch size to $2$ to avoid memory errors and recalibrated the learning rate to $0.0025$ using the formula provided in the Detectron2 documentation\footnote{\url{https://github.com/facebookresearch/detectron2/issues/1128}}. This adjustment stabilized the training process and ensured consistent convergence across all architectures, including our proposed one.

\paragraph{Hyperparameter Choices}
In our experiments, we utilized a scaling factor, $\gamma$, of $20$ for the cosine-similarity-based classifier. This decision was guided by the findings of \cite{wang2020frustratingly}, which demonstrated that a scaling factor of $20$ achieves an effective balance between reducing intra-class variance and maintaining high detection accuracy for novel classes. This choice minimizes the drop in detection accuracy for base classes, a critical consideration in few-shot scenarios with limited training examples.

\paragraph{Benchmarks and Metrics}
We benchmarked our framework against several state-of-the-art architectures, including TFA w/cos, A-RPN, DeFRCN, FRCN-ft, and CD-FSOD. To ensure fairness and consistency, we utilized 11\% of the total data available in BDTSD \cite{ashik2022recognizing} as the query set, as suggested by \cite{wang2020frustratingly}. We employed COCO-style average precision (AP) as our evaluation metric, measured over thresholds ranging from $0.5$ to $0.95$ with a step size of $0.05$. 

\paragraph{Reproducibility}
To maintain transparency and facilitate future research, all code related to our implementation will be made publicly available on GitHub upon acceptance of this manuscript.

\subsection{Model Evaluation and Analysis}

\subsubsection{Performance comparison with the State-of-the-Art in Few-Shot Traffic Sign Detection}

As shown in \autoref{tab:comparisonSOTA}, the performance of our proposed architecture was benchmarked against several state-of-the-art FSOD models on BDTSD dataset across 1-shot, 3-shot, 5-shot, and 10-shot scenarios.

We evaluated DETR-ft and YOLOv8-ft, two advanced architectures representing single-stage detectors on the BDTSD dataset. However, these models underperformed in few-shot scenarios. DETR-ft achieved an average mAP of only $7.74$, struggling particularly in the lower-shot conditions due to its heavy reliance on large datasets and prolonged training to optimize its complex attention mechanisms. Similarly, YOLOv8-ft attained an average mAP of $27.83$, surpassing CD-FSOD but failing to compete with FUSED-Net. While YOLOv8-ft demonstrated reasonable performance, its lack of a dedicated region proposal network, crucial for localizing objects in data-scarce environments, limits its effectiveness in FSOD tasks. Hence, we moved on to two-stage detectors.

Despite a moderate showing, TFA w/cos achieved the lowest average mAP of $7.03$. Even though its performance improved incrementally with the number of shots, it remained significantly behind other models, particularly in the higher-shot settings, underscoring its limited scalability in few-shot scenarios. This may be attributed to the frozen backbone of the architecture failing to generalize in novel settings of BDTSD.

A-RPN demonstrated better overall performance than TFA w/cos, with a notable improvement in average mAP to $10.02$. While it exhibited steady gains as the shot count increased, peaking at $12.22$ in the 10-shot scenario, it still fell short compared to more recent approaches. We hypothesize that the performance of A-RPN is highly contingent on the quality of the target images, which can be inconsistent in real-world scenarios like traffic sign detection. This inconsistency likely contributed to its relatively poor performance on the BDTSD, highlighting the challenges of applying attention mechanisms in the aforementioned scenario.

\begin{table*}[tb]
    \centering
    \caption{Comparison of mAP Performance across Different Categories of the Bangladeshi Traffic Sign Detection Dataset (BDTSD) under 1-shot, 3-shot, 5-shot, and 10-shot Scenarios. The categories are sorted alphabetically.}
    \label{tab:perClassPerformance}
    \begin{tabular}{L{0.25\textwidth} *{5}{R{0.065\textwidth}}}
        \toprule
        \textbf{Category} & \textbf{1-shot} &  \textbf{3-shot} &  \textbf{5-shot} &  \textbf{10-shot} & $\mathbf{\mu}$\\
        \midrule
        Cross Roads & 16.94 & 49.84 & 35.32 & 63.11 & 41.30\\
Narrow Bridge & 32.56 & 38.15 & 53.64 & 77.73 & 50.52\\
No Overtaking & 15.55 & 58.39 & 56.99 & 56.83 & 46.94\\
No Use Of Horn & 28.28 & 35.09 & 55.22 & 41.14 & 39.93\\
Pedestrain Crossing & 38.49 & 41.08 & 52.90 & 61.93 & 48.60\\
Road Hump & 32.17 & 44.95 & 45.25 & 57.69 & 45.01\\
School & 37.22 & 45.81 & 53.29 & 59.89 & 49.05\\
Sharp Bend To The Left & 28.23 & 40.05 & 36.66 & 42.96 & 36.97\\
Sharp Bend To The Right & 19.44 & 19.33 & 14.73 & 33.20 & 21.67\\
Side Road Left & 25.44 & 38.21 & 29.50 & 49.78 & 35.73\\
Side Road Right & 36.33 & 46.20 & 53.29 & 48.86 & 46.17\\
Speed Limit 40 km/h & 33.23 & 54.35 & 55.49 & 52.50 & 48.89\\
Speed Limit 60 km/h & 54.13 & 33.29 & 67.55 & 73.00 & 56.99\\
Speed Limit 80 km/h & 27.04 & 54.20 & 56.95 & 64.63 & 50.70\\
U Turn & 4.34 & 39.00 & 32.22 & 31.88 & 26.86\\
\bottomrule
    \end{tabular}
\end{table*}

DeFRCN outperformed both TFA w/cos and A-RPN, with a significant leap in average mAP to $23.15$. This model showed remarkable improvements in higher-shot scenarios, particularly excelling in the 5-shot and 10-shot conditions, reflecting its effectiveness in leveraging additional shot information. However, DeFRCN's reliance on certain frozen modules, particularly in the RPN, limits its ability to improve classification scores when working with low-quality region proposals. This limitation may have hindered its ability to perform better in BDTSD, where due to the complexity of the road images, it is difficult to generate good region proposals.

With an average mAP of $25.81$, FRCN-ft delivered strong performance, particularly in higher-shot scenarios. Its mAP soared from $11.36$ in the 1-shot scenario to $42.12$ in the 10-shot scenario, indicating robust adaptation to increased shot availability. Its simplicity and effectiveness underscore the importance of adaptable yet straightforward architectures in FSOD, especially in domains where the object instances, such as traffic signs, exhibit relatively consistent features. The significant performance gains of FRCN-ft in higher-shot scenarios suggest that a well-calibrated fine-tuning process can match, and sometimes exceed, the performance of more complex models like CD-FSOD, especially when computational efficiency is a concern.

Among the compared models, CD-FSOD achieved the second-highest performance, with an average mAP of $26.45$. Its strengths were most evident in the 5-shot and 10-shot scenarios, where it attained mAPs of $31.02$ and $43.02$, respectively, showing its capability to scale effectively with more training data. However, as mentioned before, the complexity of the CD-FSOD approach, with its reliance on a teacher-student distillation framework, may be over-engineered for specific tasks like traffic sign detection, where objects have relatively consistent shapes and colors. This observation inspired our approach, which simplifies the FSOD process by leveraging a pseudo-support set that facilitates \modelName\, a model with 60 million parameters, to perform better than CD-FSOD which requires 82 million. Hence, this simplification enhances detection performance while maintaining computational efficiency.

Our proposed architecture, integrating the strengths of the state-of-the-art approaches while addressing their limitation, consistently outperformed all other models across every shot scenario, achieving the highest average mAP $42.81$. Specifically, the inclusion of a cosine-similarity-based classifier, inspired by TFA w/cos, enhances embedding normalization and comparability across instances, significantly improving performance in few-shot scenarios. Additionally, we incorporate domain adaptation strategies to ensure robust generalization to the target domain. However, unlike the complex methodologies of CD-FSOD, we achieve this through a simpler, more efficient process that is particularly well-suited to the consistent nature of traffic sign detection. Our pseudo-support set technique enriches the training process by introducing necessary variability, crucial in few-shot learning, and ensures that the model is well-calibrated to the specific characteristics of BDTSD. 

In conclusion, our model \modelName\ not only achieves superior performance in terms of mAP but also exemplifies a balanced approach to FSOD, combining simplicity and robustness. Our performance gain underscores the importance of a holistic strategy that addresses both data variability through pseudo-support sets and domain specificity via adaptation techniques, positioning \modelName\ as a leading solution in few-shot traffic sign detection.

\subsubsection{Per-Category Performance Analysis}
To provide a more granular evaluation of FUSED-Net, we analyze its performance across individual traffic sign categories in the BDTSD dataset. \autoref{tab:perClassPerformance} reports the mAP values for each of the 15 categories under 1-shot, 3-shot, 5-shot, and 10-shot scenarios.

The analysis reveals notable variablity in performance across different categories. For instance, the model achieves high average mAP for ``Speed Limit 60 km/h'' (56.99) and ``Narrow Bridge'' (50.52), indicating the effective detection in these scenarios. However, the performance for certain categories like ``U Turn'' (26.68) and ``Sharp Bend to the Right'' (21.67) is relatively lower, suggesting that these categories present greater challenges for few-shot detection. The underperformance in the ``U Turn" and ``Sharp Bend to the Right" categories can be attributed to a combination of factors. These include geometric complexity and ambiguity, where subtle distinctions in directional arrows are challenging to capture with limited training data, leading to potential misclassifications. Additionally, the minimalist visual features of these signs, often lacking text or intricate patterns, make them harder to distinguish, especially under conditions like occlusion or poor lighting. Few-shot models face inherent challenges in fine-grained detection tasks, as limited samples restrict their ability to generalize nuanced differences, further compounded by category distribution imbalances. Lastly, these signs often depend on contextual cues from the road environment, which are difficult to incorporate effectively with minimal training data.

The analysis underscores that while FUSED-Net demonstrates robust performance in most categories, it also highlights the impact of category-specific characteristics on detection accuracy. This evaluation provides further insight into the strengths and potential areas of improvement for FUSED-Net, particularly in challenging traffic sign categories.

\subsubsection{Statistical Significance of the Performance}
\begin{figure}[bt]
    \centering
    \includegraphics[width=0.85\textwidth]{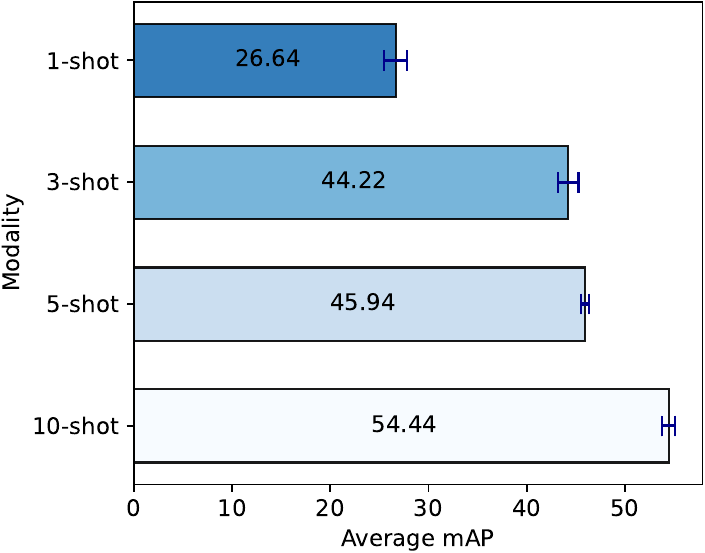}
    \caption{Average mAP across different shot scenarios with 95\% confidence intervals. Each bar represents the mean mAP for a given shot scenario, with error bars indicating the range of variability.}
    \label{fig:statSig}
\end{figure}

Considering the substantial performance gap of FUSED-Net over other SOTA architectures, we conducted multiple independent trials for each of the 1-shot, 3-shot, 5-shot, and 10-shot scenarios to ensure the reproducibility and validity of our findings. After performing the trials, we calculated the mean and $95\%$ confidence intervals (CI95) for the mAP values across different few-shot scenarios. The values are shown in \autoref{fig:statSig}. The results demonstrate a consistent performance advantage of FUSED-Net compared to SOTA methods. The relatively narrow confidence intervals affirm the stability and reliability of the proposed architecture under various few-shot scenarios.

To provide a more comprehensive evaluation, we also analyzed the precision, recall, and F1 scores achieved by FUSED-Net as summarized in \autoref{tab:preRecF1}. The precision and recall values indicate that the model maintains an effective balance between detecting relevant instances (recall) and minimizing false positives (precision) across all shot scenarios. Notably, recall exhibits slightly higher values than precision, suggesting that FUSED-Net prioritizes capturing more potential traffic sign instances, even if it occasionally predicts additional false positives. The F1-score, representing the harmonic mean of precision and recall,  improves consistently as the number of shots increases, demonstrating the model's ability to generalize better with additional data.

\begin{table}[tb]
    \centering
    \caption{Average precision, recall, and F1 scores of FUSED-Net under 1-shot, 3-shot, 5-shot, and 10-shot scenarios.}
    \label{tab:preRecF1}
    \begin{tabular}{l r r r r}
        \toprule
        \textbf{Modality} & \textbf{1-shot} & \textbf{3-shot} & \textbf{5-shot} & \textbf{10-shot}\\
        \midrule
        Precision & 0.29 & 0.43 & 0.47 & 0.54\\
        Recall & 0.46 & 0.57 & 0.58 & 0.63\\
        F1-Score & 0.35 & 0.49 & 0.52 & 0.58\\
        \bottomrule
    \end{tabular}
\end{table}

To further validate the significance of these findings, we performed a statistical significance analysis using paired t-tests \cite{helmert1876genuigkeit} to compare FUSED-Net's performance with the next-best-performing SOTA method (CD-FSOD). Across all scenarios, a t-statistic of 5.73 and a p-value of 0.0106 confirm that the observed performance differences are statistically significant.

The narrow confidence intervals, coupled with the precision, recall, and F1-score analysis and the statistical significance of our findings, demonstrate the high reliability of FUSED-Net’s performance. These results instill confidence in the practical applicability of our approach, particularly in real-world scenarios where data availability is often limited.

\subsection{Model Insights and Behavior}

\subsubsection{Ablation Study}

\begin{table*}[tb]
    \centering
    \caption{Ablation study on the Bangladeshi Traffic Sign Detection Dataset (BDTSD) showing the impact of incorporating various modifications into the proposed pipeline. Here, the columns with $\Updelta$ indicate the increase in the mAP relative to the baseline model, demonstrating the contribution of each component to the overall performance.}
    \label{tab:ablation}
    \begin{tabular}{C{1cm}C{0.7cm}C{0.7cm}C{0.7cm} | C{1.2cm}C{0.5cm} C{1.2cm}C{0.5cm} C{1.2cm}C{0.5cm} C{1.2cm}C{0.5cm}}
    \toprule
        \textbf{UP} & \textbf{PSS} & \textbf{EN}  & \textbf{DA} & \textbf{1-shot} & $\mathbf{\Updelta}$ & \textbf{3-shot} & $\mathbf{\Updelta}$ & \textbf{5-shot} & $\mathbf{\Updelta}$& \textbf{10-shot} & $\mathbf{\Updelta}$\\
    \midrule
    \xmark      & \xmark & \xmark & \xmark & 10.4 & - & 18.8 & - & 26.6 & - & 38.6 & -\\
    \checkmark  & \xmark & \xmark & \xmark & 11.2 & 0.8 & 20.2 & 1.3 & 28.7 & 2.1 & 40.0 & 1.5\\
    \checkmark  & \checkmark & \xmark & \xmark & 11.4 & 1.0 & 21.2 & 2.4 & 29.7 & 3.1 & 42.2 & 3.7\\
    \checkmark  & \checkmark  & \checkmark & \xmark & 13.0 & 2.5 & 22.2 & 3.3 & 30.6 & 4.0 & 43.1 & 4.6\\
    \checkmark  & \checkmark  & \checkmark  & \checkmark  & 26.6 & 16.2 & 44.2 & 25.4 & 45.9 & 19.3 & 54.4 & 15.9\\
    \bottomrule
    \multicolumn{12}{L{15cm}}{(UP = Unfrozen Parameters, PSS = Pseudo Support Set, EN = Embedding Normalization, and DA = Domain Adaptation)} 
    \end{tabular}
\end{table*}

To assess the contribution of the individual components of \modelName, we conducted an ablation study. The study was performed on BDTSD across 1-shot, 3-shot, 5-shot, and 10-shot scenarios. Our model integrates four key modifications: unfrozen parameters, pseudo-support set, embedding normalization, and domain adaptation. The results, presented in \autoref{tab:ablation}, demonstrate the incremental improvements each modification contributes to the overall system performance, measured by mAP.

\paragraph{Unfrozen Parameters}
The decision to unfreeze all layers of our architecture during fine-tuning stage played a pivotal role in improving cross-domain detection accuracy. Unlike conventional practices where certain layers are frozen to retain pre-trained knowledge, our approach allows every layer---from the Backbone to the Region Proposal Network, Feature Pyramid Network, and fully connected layers---to adjust its weights. This flexibility is helpful in scenarios where the target domain (e.g., Bangladeshi Traffic Signs) is significantly different from the base domain (e.g., general object detection datasets). As a result, the model can learn new, domain-specific features that are essential for detecting traffic signs under diverse conditions.

While the ablation study reveals that unfreezing parameters alone yields a moderate improvement of $0.8\%$ mAP in the 1-shot scenario, this result aligns with findings from prior research \cite{xiong2023cd}, which demonstrated that unfrozen parameter yields varying degrees of improvement across different FSOD architectures (as illustrated in \autoref{fig:impactOfUnfrozenParameters}). For example, DeFRCN, a SOTA FSOD architecture known for its generalizability, achieved only a 1.2\% improvement with unfrozen parameters. Similarly, our proposed architecture, designed for high generalizability, exhibits smaller yet meaningful gains, particularly under the highly challenging 1-shot setting.

Further, the impact of unfrozen parameters becomes more apparent in the 5-shot scenario, where a 2.1\% improvement in mAP is observed. This trend suggests that while the benefits of unfreezing are modest in extremely low-data regimes, they become increasingly effective as more training data becomes available. Overall, these results demonstrate that unfreezing parameters contributes meaningfully to FUSED-Net's adaptability and performance, particularly as the difficulty of the task or the availability of data increases.

\paragraph{Pseudo Support Set}
To overcome the challenge of limited training data inherent in FSL, we introduced a pseudo-support set through data augmentation. By applying random color jitter to the original support set, we created a richer, more diverse dataset that better represents the variability encountered in real-world scenarios. This approach proved to be particularly effective, as evident from the performance gains. For example, the addition of the pseudo-support set increased the mAP by $1.0\%$ in the 1-shot scenario and by a more pronounced $3.7\%$ in the 10-shot scenario. The augmented support set allowed the model to generalize more effectively, improving its ability to detect traffic signs despite the limited number of examples available for training.

To validate our choice of Color Jittering, we conducted additional experiments comparing the effects of various augmentation strategies such as random cropping and horizontal flipping --- both individually and in combination with Color Jitter. The results demonstrated that employing Color Jitter alone consistently produced the highest mAP scores across all shot scenarios. These findings reinforce that the simplicity and effectiveness of Color Jitter make it the best choice for augmenting the PSS.

\paragraph{Embedding Normalization}
Incorporating embedding normalization further refined our model's ability to handle high intra-class variance. By standardizing the feature representations of individual instances, our approach reduces the impact of outliers and ensures that features are comparable across different instances. This normalization is enabled by a cosine similarity-based classifier, which focuses on the angular relationships between feature vectors, thereby reducing the influence of magnitude differences. The results shown in \autoref{tab:ablation} highlight the significant improvement this modification brings, particularly in 10-shot scenarios, where mAP increased by $4.6\%$ compared to the baseline. This demonstrates the effect of embedding normalization in enhancing the model's robustness and generalization capabilities.

\begin{figure*}[tb]
\centering
\subfloat[Detection and classification of a distant, blurry traffic sign. Despite the challenging conditions, the model accurately identified the sign.\label{subfig:hardDistant}]{
\includegraphics[width=.45\textwidth]{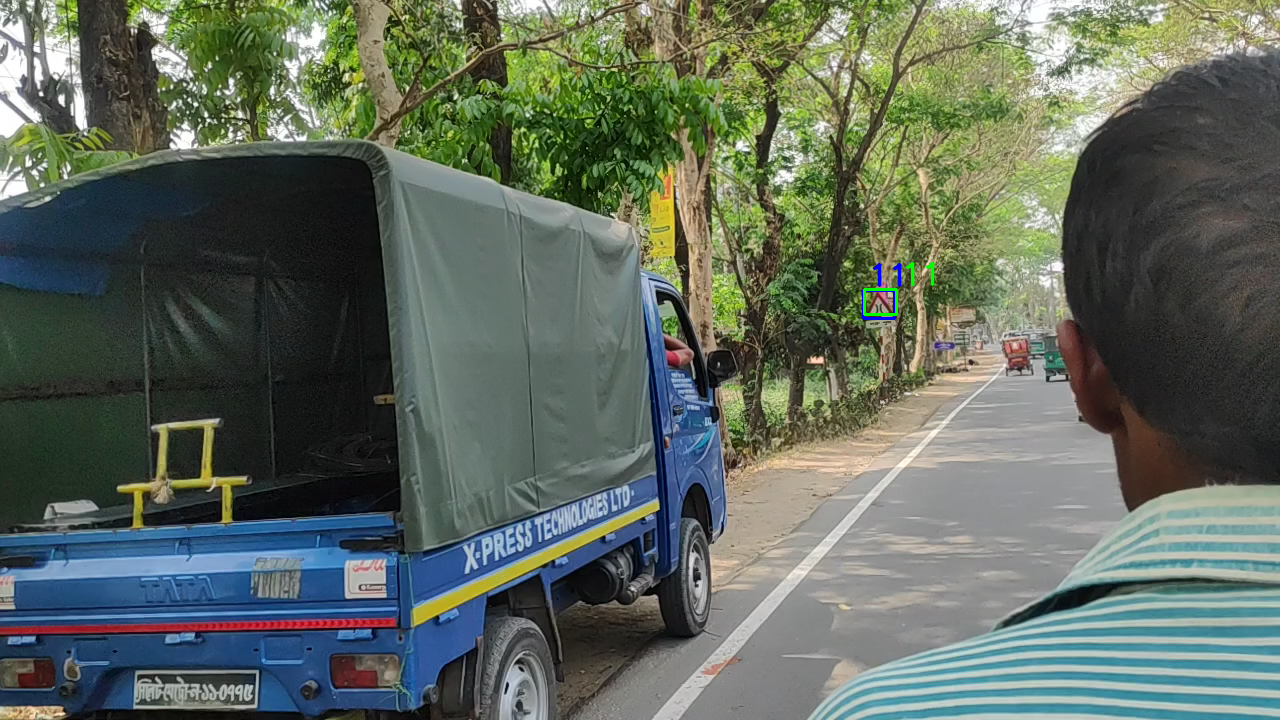}   
}
\hspace{1em}
\subfloat[Detection and classification of two traffic signs at different distances. The model correctly predicted both the closer and further signs, proving its effectiveness in identifying multiple signs in varying proximities.\label{subfig:hardMultiple}]{
\includegraphics[width=.45\textwidth]{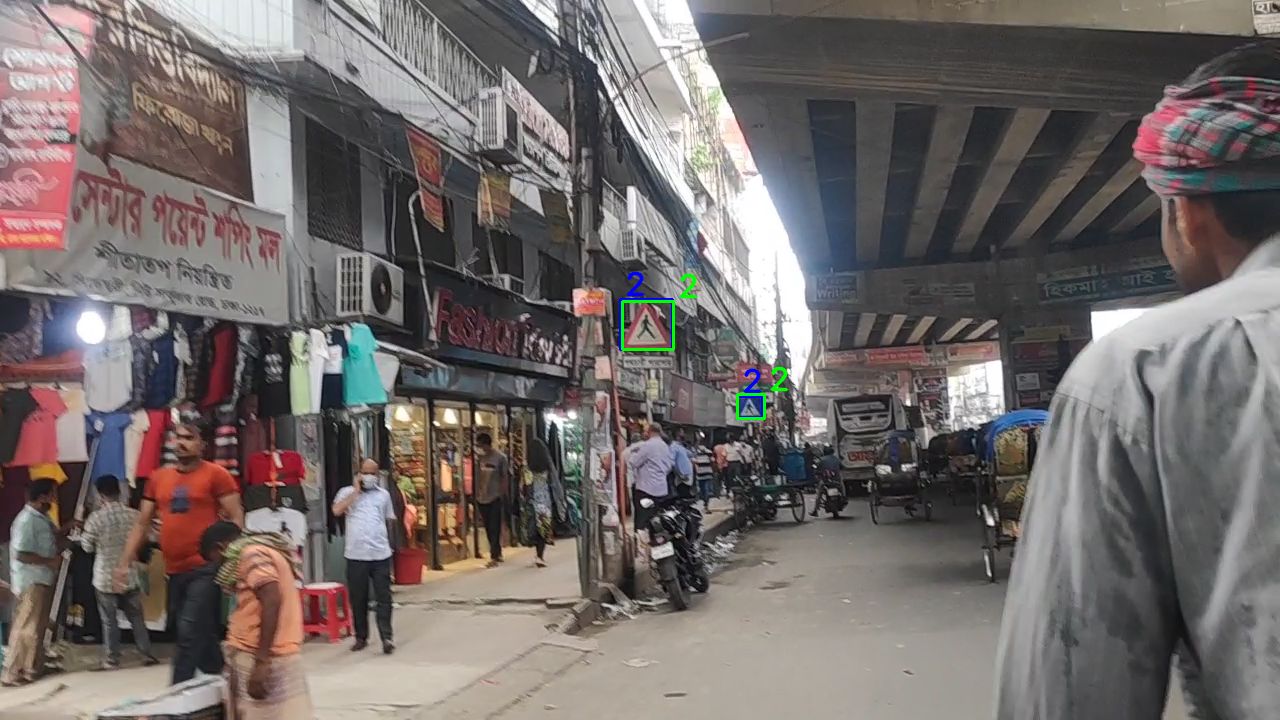}
}
\caption{Qualitative examples of successful traffic sign detection and classification by our model \modelName\ on the BDTSD. Here, the green bounding box and labels are predictions from \modelName\ and the blue ones are ground truths.}
\label{fig:visualHard}
\end{figure*}

\paragraph{Domain Adaptation}
Domain adaptation had the most significant impact on our model's performance. By training the model on traffic signs from diverse countries, we reduced the distributional variance between the source and target domains. This strategy was crucial in enabling the model to generalize effectively across different environments. The results in \autoref{tab:ablation} underscore the effectiveness of domain adaptation, showing a dramatic increase in mAP across all shot scenarios. Even in the 1-shot scenario, the mAP jumps by $16.2\%$ compared to the baseline. It remains substantial across higher-shot scenarios, demonstrating the effectiveness of this technique in addressing the challenges posed by domain discrepancies.

\subsubsection{Qualitative Analysis}
\begin{figure}[t]
\centering
\includegraphics[width=.98\textwidth]{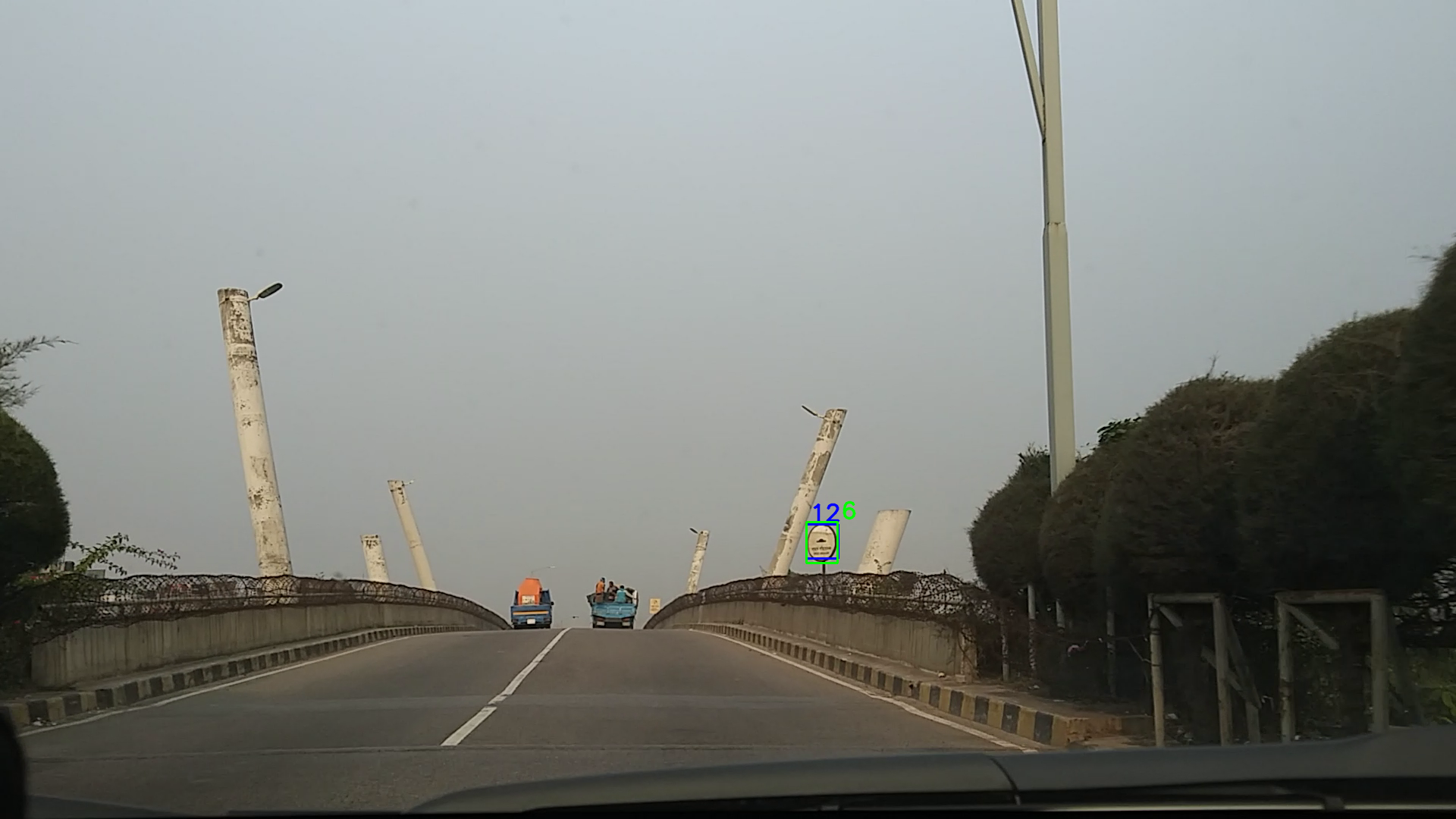}
\caption{An example of successful traffic sign detection but incorrect classification by our model on the BDTSD. The sign is extremely distant and features a complex symbol, making it challenging to classify accurately. Here, the green bounding box and label are predictions from our model and the blue ones are the ground truths.}
\label{fig:visualWrong}
\end{figure}

For the qualitative analysis, we evaluated our model's performance by fine-tuning it on the MTSD dataset and testing it with a 5-shot scenario from BDTSD. In Figs. \ref{fig:visualHard} and \ref{fig:visualWrong}, the green bounding boxes represent detections made by our model, while the blue bounding boxes denote the ground truth annotations. The labels above the blue boxes indicate the actual class labels, whereas the labels above the green boxes indicate the class labels predicted by our model. Note that, only detected boxes with confidence scores of $0.5$ or higher are displayed.

As demonstrated in \autoref{subfig:hardDistant}, our model effectively detects traffic signs that are blurry and located at significant distances from the camera, a challenging scenario that often leads to detection failures in other models. Additionally, \autoref{subfig:hardMultiple} shows that our model can accurately detect and classify multiple traffic signs within a single image, even when they are positioned at varying distances and embedded in complex backgrounds. This ability to handle both single and multiple instances under challenging conditions underscores the robustness of our approach in real-world scenarios.

However, our analysis also reveals areas for improvement. As shown in \autoref{fig:visualWrong}, while the model successfully detects traffic signs that are extremely distant and blurred, it sometimes fails to correctly classify them. This misclassification likely stems from the difficulty of discerning complex symbols on distant signs, a challenge even for human observers. This limitation suggests that although our model is effective in detection, the performance of the classifier could be enhanced by refining its ability to interpret subtle visual cues in challenging contexts. Future work could focus on improving the sensitivity of the classifier to these complex visual patterns, especially in low-visibility conditions, to further enhance the overall performance of the model.

\subsection{Performance Analysis on Cross-Domain Few-Shot Object Detection (CD-FSOD) Benchmark}

To thoroughly assess the cross-domain performance of our proposed architecture, we conducted extensive experiments on the CD-FSOD benchmark, which evaluates models using base datasets from one domain (MS COCO) and novel datasets from distinct domains (ArTaxOr, UODD, DIOR). The results of these experiments, as shown in \autoref{tab:ComparisonWithOriginalCdfsodDataset}, offer insights into the efficacy of our approach compared to the state-of-the-art (SOTA) FSOD models.

\begin{table*}[tb]
    \caption{Comparison of mAP Performance Across Various Few-Shot Object Detection Models on the Cross-domain Few-shot Object Detection (CD-FSOD) Benchmark under 1-Shot, 5-Shot, and 10-Shot Scenarios across Three Datasets: Arthropod Taxonomy Orders Object Detection Dataset (ArTaxOr), Underwater Object Detection Dataset (UODD), and Dataset for Object Detection in Aerial Images (DIOR). Models are sorted by the average mAP ($\mu$) across all shots and datasets, with the best results highlighted in boldface.}
    \label{tab:ComparisonWithOriginalCdfsodDataset}
    \begin{tabular}{L{0.16\textwidth} *{2}{R{0.057\textwidth}} R{0.065\textwidth} *{2}{R{0.057\textwidth}} R{0.065\textwidth} *{2}{R{0.057\textwidth}} R{0.065\textwidth} R{0.04\textwidth}}
        \toprule
         & \multicolumn{3}{c}{\textbf{ArTaxOr}} & \multicolumn{3}{c}{\textbf{UODD}} & \multicolumn{3}{c}{\textbf{DIOR}}\\
        \cmidrule{2-10}
         \textbf{Model} & \textbf{1-shot} & \textbf{5-shot} & \textbf{10-shot} & \textbf{1-shot} & \textbf{5-shot} & \textbf{10-shot} & \textbf{1-shot} & \textbf{5-shot} & \textbf{10-shot} & $\mathbf{\mu}$\\
         \midrule
         TFA w/cos \cite{wang2020frustratingly}  & 0.8 & 3.8 & 7.1 & 2.7 & 6.5 & 7.2 & 3.9 & 10.3 & 13.4 & 6.2\\
        A-RPN \cite{fan2020few} & 1.4 & 5.0 & 9.7 & 2.3 & 6.1 & 9.2 & 6.3 & 14.4 & 17.9 & 8.0\\
        FSCE \cite{sun2021fsce} & 1.8 & 5.9 & 10.8 & 2.8 & 6.7 & 8.4 & 5.6 & 14.9 & 18.3 & 8.4\\
        Meta-RCNN \cite{han2022meta} & 1.9 & 6.1 & 10.3 & 2.8 & 6.7 & 9.9 & 5.5 & 15.2 & 18.8 & 8.6\\
        H-GCN \cite{han2021query} & 2.0 & 6.3 & 10.9 & 4.5 & 6.2 & 9.4 & 6.0 & 15.4 & 18.7 & 8.8\\
        DeFRCN \cite{qiao2021defrcn} & 2.9 & 8.8 & 14.5 & 3.7 & 8.9 & 10.7 & 8.0 & 17.7 & 20.7 & 10.7\\
        FRCN-ft \cite{ren2015faster} & 3.4 & 9.3 & 15.2 & 4.1 & 9.2 & 12.3 & 8.4 & 18.3 & 21.2 & 11.3\\
        CD-FSOD \cite{xiong2023cd} & \textbf{5.1} & 12.5 & 18.1 & \textbf{5.9} & 12.2 & \textbf{14.5} & 10.5 & 19.1 & \textbf{26.5} & 13.8\\
        \modelName\ (Ours) & 3.6 & \textbf{16.4} & \textbf{23.5} & 4.4 & \textbf{13.6} & 12.7 & \textbf{11.4} & \textbf{19.2} & 23.3 & \textbf{14.2}\\
\bottomrule
    \end{tabular}
\end{table*}

As discussed before, the traditional FSOD benchmarks, such as MS COCO and Pascal VOC, typically involve dividing data into base and novel sets drawn from the same domain. However, this does not reflect real-world applications where domain shifts are common. For instance, a model trained on natural images might need to detect objects in aerial or optical remote sensing images---domains with significantly different characteristics. Models fine-tuned within the same domain may perform well under such circumstances but often struggle when applied across domains. To identify whether it can excel in cross-domain settings, our proposed model, while originally intended for few-shot traffic sign detection, was subjected to rigorous testing on the CD-FSOD benchmark to evaluate its versatility. 

While our model achieves the highest average mAP across all datasets and shot settings ($14.2$), outperforming the leading SOTA models, including CD-FSOD, it exhibits inconsistencies in the UODD dataset at the 10-shot scenario ($12.7\%$ vs CD-FSOD's $14.5\%$). This discrepancy likely stems from differences between traffic sign images and underwater object images. Underwater scenes often feature clear, uniform backgrounds (e.g., blue ocean water) and slightly blurry objects, whereas traffic scenes contain highly cluttered and noisy backgrounds.

The Pseudo Support Set technique used in FUSED-Net is highly effective for challenging scenarios like traffic signs with complex surroundings, enabling it to detect such objects robustly. However, in simpler environments like underwater scenes, this method may inadvertently cause overfitting, which limits its adaptability. Despite these domain-specific challenges, our model outperforms SOTA models in overall cross-domain performance and consistently excels in 5-shot scenarios, emphasizing its utility in diverse applications.

SOTA architectures, such as TFA w/cos, A-RPN, FSCE, Metha-RCNN, H-GCN, and DeFRCN, involve freezing certain components during fine-tuning, which may stabilize the model within the same domain but often leads to poor performance in cross-domain settings. The data clearly shows that models with frozen components struggle with significant domain shifts, achieving lower mAPs across the board. For instance, TFA w/cos only achieved a $7.2$ average mAP, reflecting its limited adaptability when faced with diverse domains.

Although FRCN-ft and CD-FSOD do not freeze any components, our model surpasses them in average performance. The performance disparity with FRCN-ft stems from its reliance on a limited number of labeled samples during fine-tuning, which hampers its ability to generalize in FSL. This scarcity leads to inadequate learning of novel classes, especially in complex domains such as the CD-FSOD benchmark, resulting in suboptimal performance. In contrast, our model utilizes a pseudo-support set technique to amplify the data available during fine-tuning by generating additional samples from labeled data. This broader data exposure allows our model to learn more robust features, enhancing generalization and reducing overfitting, leading to superior performance.

Our proposed \modelName\ model even surpasses the average performance of CD-FSOD, which was designed specifically for cross-domain scenarios. This success can be attributed to \modelName's straightforward yet effective training strategy. Unlike CD-FSOD, which employs a complex two-model system with adaptive learning based on augmentations, our approach uses a single model trained with a pseudo-support set. This method not only simplifies the training process but also enhances generalization by effectively increasing the data supply during fine-tuning.

In conclusion, while FUSED-Net demonstrates slight domain-specific limitations in simpler datasets like UODD, the overall results firmly establish it as a versatile and robust architecture for cross-domain FSOD. This achievement, coupled with its simplicity and robustness, positions our approach as a highly effective solution for real-world FSOD, even in scenarios involving significant domain shifts.

\section{Conclusion}\label{sec:conclusion}

In this work, we presented \modelName, a novel Few-Shot Traffic Sign Detector, designed to address the challenges of traffic sign detection in scenarios with limited labeled data. Our model integrates domain adaptation and a pseudo-support set to significantly enhance the detection accuracy of traffic signs, where state-of-the-art methods often falter. Both approaches can be seamlessly incorporated into existing detection architectures without disrupting the training process, offering practical solutions to the data scarcity problem in traffic sign detection in remote areas. We also examined the limitations of freezing model components, a common practice in current SOTA architectures, which we found to be detrimental in few-shot detection tasks. By allowing for greater adaptability through unfrozen parameters, combined with embedding normalization, \modelName\ demonstrates superior generalization capabilities across diverse domains. Our work not only advances the field of few-shot traffic sign detection but also highlights the importance of adaptable model components and domain-aware techniques in overcoming the challenges of few-shot learning. We hope this work inspires further innovation and exploration in few-shot learning, particularly in specialized domains like traffic sign detection, where data availability is often limited.

\bibliographystyle{IEEEtran}
\bibliography{access-Ref}

\begin{IEEEbiography}
[{\includegraphics[width=1in,height=1.25in,clip,keepaspectratio]{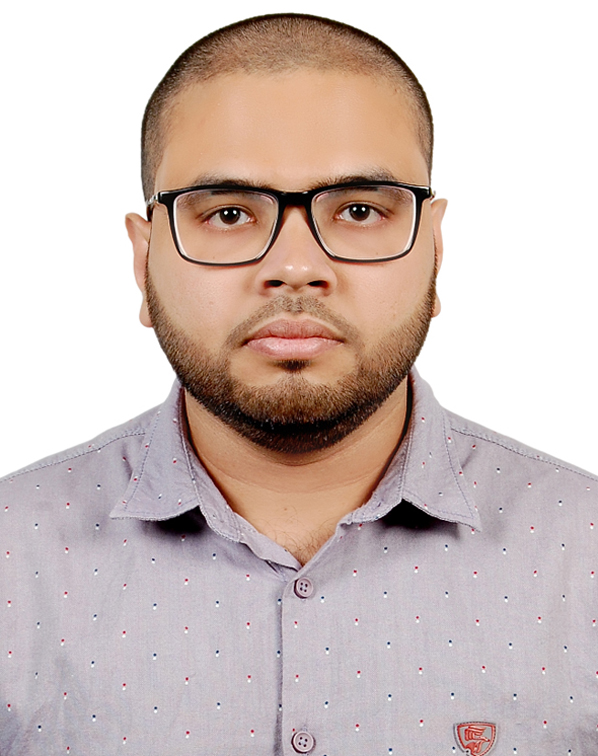}}]{Md. Atiqur Rahman} completed his B.Sc. Engg. in Computer Science and Engineering (CSE) from the Islamic University of Technology (IUT) in 2023 and is currently pursuing his M.Sc. Engg. in CSE from the same institution.

He began his career as a Lecturer in the Department of Computer Science and Engineering at United International University in 2023. Shortly after, he joined the Islamic University of Technology as a Lecturer in the same department. His research interests include deep learning-based computer vision techniques in few-shot object detection and image generation. 
\end{IEEEbiography}

\begin{IEEEbiography}[{\includegraphics[width=1in,height=1.25in,clip,keepaspectratio]{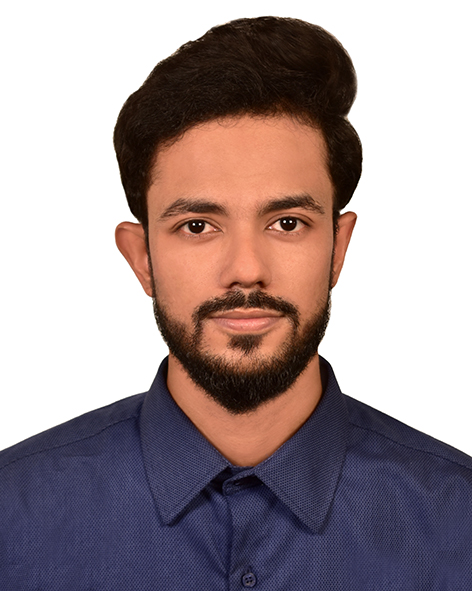}}]{Nahian Ibn Asad} completed his B.Sc. Engg. degree in Computer Science and Engineering (CSE) from Islamic University of Technology (IUT) in 2023.

He is currently working as a Software Engineer in the Machine Learning team at TigerIT Bangladesh Ltd. He has worked on several high-impact projects for the Government of Bangladesh, including the development of the Automatic Number Plate Recognition (ANPR) system for the Bangladesh Road Transport Authority (BRTA) and the Face Anti-Spoofing system for the Bangladesh Election Commission. At present, he is working on an international project in collaboration with Poland. His research interests lie in areas like object detection, image processing, contextual breaches, and the robustness of language models through deep learning, computer vision, and natural language processing techniques.
\end{IEEEbiography}

\begin{IEEEbiography}
 [{\includegraphics[width=1in,height=1.25in,clip,keepaspectratio]{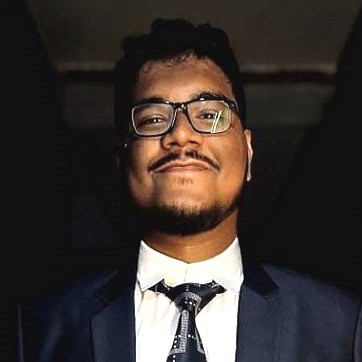}}]{Md. Mushfiqul Haque Omi} earned his B.Sc. in Computer Science and Engineering (CSE) from the Islamic University of Technology (IUT) in 2023. He is currently pursuing an M.Sc. in CSE at Dhaka University. 

 His academic career as a Lecturer began in 2023 in the Department of CSE at Bangladesh University of Business and Technology (BUBT). In 2024, he joined United International University (UIU) as a Lecturer in the same department. His research focuses on deep learning-based computer vision techniques for few-shot object detection, as well as computer networking and security.
 \end{IEEEbiography}

\begin{IEEEbiography}[{\includegraphics[width=1in,height=1.25in,clip,keepaspectratio]{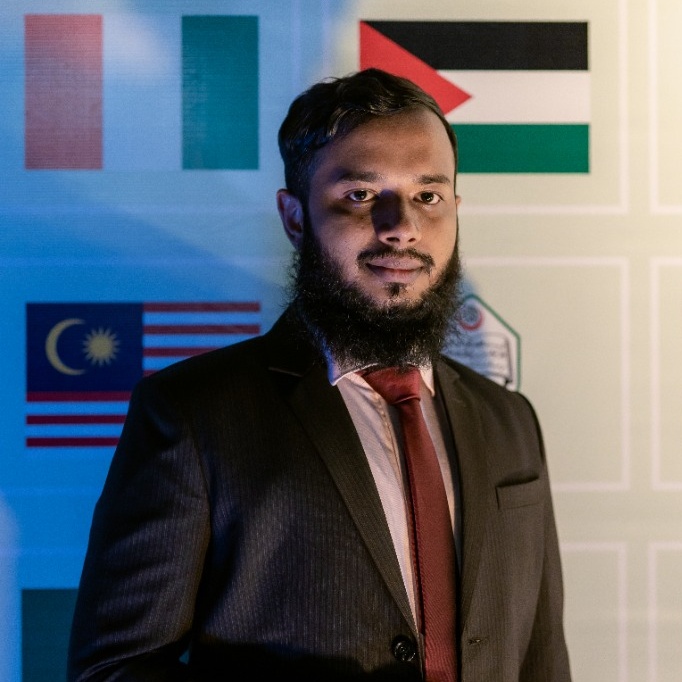}}]{Md. Bakhtiar Hasan} completed his B.Sc. Engg. and M.Sc. Engg. degree in Computer Science and Engineering (CSE) from Islamic University of Technology (IUT) in 2018 and 2022, respectively.

He started his career as a Lecturer in the Department of Computer Science and Engineering, IUT, and was promoted to Assistant Professor in 2022. His research interest includes the use of deep learning and computer vision techniques in human biometrics and smart agriculture.

Mr. Hasan received the Huawei Seeds for the Future scholarship in 2018. He was awarded IUT Gold Medal in recognition of his outstanding performance in the pursuit of the B.Sc. Engg. in CSE degree in 2018.
\end{IEEEbiography}

\begin{IEEEbiography}[{\includegraphics[width=1in,height=1.25in,clip,keepaspectratio]{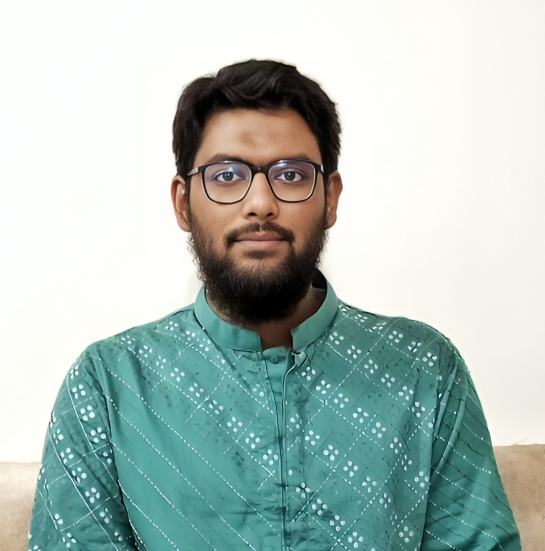}}]{Sabbir Ahmed}
is an Assistant Professor in the CSE Department at the Islamic University of Technology (IUT). He received a M.Sc. and B.Sc. (Gold medalist) in CSE from IUT, respectively, in 2022 and 2017. He is a member of the Computer Vision Research Group (CVLab) and his current research is improving few-shot learning algorithms for computer vision tasks. Besides, he has worked with different applications of deep learning in the field of Computer Vision and NLP, including leaf disease classification, gait analysis, traffic sign detection, depression assessment from social media posts, abstractive text summarization, etc. He has been a member of IEEE since 2023.
\end{IEEEbiography}

\begin{IEEEbiography}[{\includegraphics[width=1in,height=1.25in,clip,keepaspectratio]{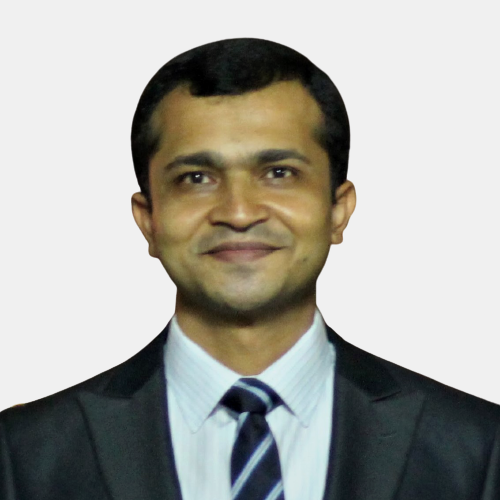}}]{Md. Hasanul Kabir} (M’17) received the B.Sc. degree in computer science and information technology from the Islamic University of Technology, Bangladesh, and the Ph.D. degree in computer engineering from Kyung Hee University, South Korea.

He is currently a Professor in the Department of Computer Science and Engineering, Islamic University of Technology. His research interests include feature extraction, visual question answering, and sign language interpretation by combining image processing, machine learning, and computer vision.
\end{IEEEbiography}

\EOD

\end{document}